\documentclass[preprint,12pt,authoryear,nopreprintline]{elsarticle}

\usepackage{amssymb}
\usepackage{url}

\usepackage{tabularx,booktabs}
\usepackage{caption} 
\usepackage{subcaption}
\usepackage{algorithm}
\usepackage{makecell}
\usepackage[noend]{algpseudocode}
\usepackage[most]{tcolorbox}
\usepackage{multirow}

\journal{Journal of Biomedical Informatics}

\begin{document}

\begin{frontmatter}



\title{Lessons from the TREC Plain Language Adaptation of Biomedical Abstracts (PLABA) track}


\author[yale]{Brian Ondov\corref{cor1}}
\author[tufts]{William Xia}
\author[nyu]{Kush Attal}
\author[jhu]{Ishita Unde}
\author[gat]{Jerry He}
\author[nlm]{Dina Demner-Fushman}

\affiliation[yale]{organization={Yale School of Medicine},
            addressline={333 Cedar Street}, 
            city={New Haven},
            postcode={06510}, 
            state={CT},
            country={USA}}

\affiliation[tufts]{organization={Tufts University},
            addressline={419 Boston Avenue}, 
            city={Medford},
            postcode={02155}, 
            state={MA},
            country={USA}}

\affiliation[nyu]{organization={NYU Grossman School of Medicine},
            addressline={550 First Avenue}, 
            city={New York},
            postcode={10016}, 
            state={NY},
            country={USA}}

\affiliation[jhu]{organization={Johns Hopkins University},
            addressline={3400 N. Charles Street}, 
            city={Baltimore},
            postcode={21218}, 
            state={MD},
            country={USA}}

\affiliation[gat]{organization={Georgia Institute of Technology},
            addressline={North Avenue}, 
            city={Atlanta},
            postcode={30332}, 
            state={GA},
            country={USA}}

\affiliation[nist]{organization={National Institute of Standards and Technology},
            addressline={100 Bureau Drive}, 
            city={Gaithersburg},
            postcode={20899}, 
            state={MD},
            country={USA}}

\affiliation[nlm]{organization={National Library of Medicine},
            addressline={8600 Rockville Pike}, 
            city={Bethesda},
            postcode={20894}, 
            state={MD},
            country={USA}}

\cortext[cor1]{brian.ondov@yale.edu}

\begin{abstract}
\textbf{Objective}:
Recent advances in language models have shown potential to adapt professional-facing biomedical literature to plain language, making it accessible to patients and caregivers. However, their unpredictability, combined with the high potential for harm in this domain, means rigorous evaluation is necessary. Our goals with this track were to stimulate research and to provide high-quality evaluation of the most promising systems.

\textbf{Methods}: We hosted the Plain Language Adaptation of Biomedical Abstracts (PLABA) track at the 2023 and 2024 Text Retrieval Conferences. Tasks included complete, sentence-level, rewriting of abstracts (Task 1) as well as identifying and replacing difficult terms (Task 2). For automatic evaluation of Task 1, we developed a four-fold set of professionally-written references. Submissions for both Tasks 1 and 2 were provided extensive manual evaluation from biomedical experts.

\textbf{Results}: Twelve teams spanning twelve countries participated in the track, with models from multilayer perceptrons to large pretrained transformers. In manual judgments of Task 1, top-performing models rivaled human levels of factual accuracy and completeness, but not simplicity or brevity. Automatic, reference-based metrics generally did not correlate well with manual judgments. In Task 2, systems struggled with identifying difficult terms and classifying how to replace them. When generating replacements, however, LLM-based systems did well in manually judged accuracy, completeness, and simplicity, though not in brevity.

\textbf{Conclusion}: 
The PLABA track showed promise for using Large Language Models to adapt biomedical literature for the general public, while also highlighting their deficiencies and the need for improved automatic benchmarking tools.
\end{abstract}

\begin{graphicalabstract}
\includegraphics[width=\textwidth]{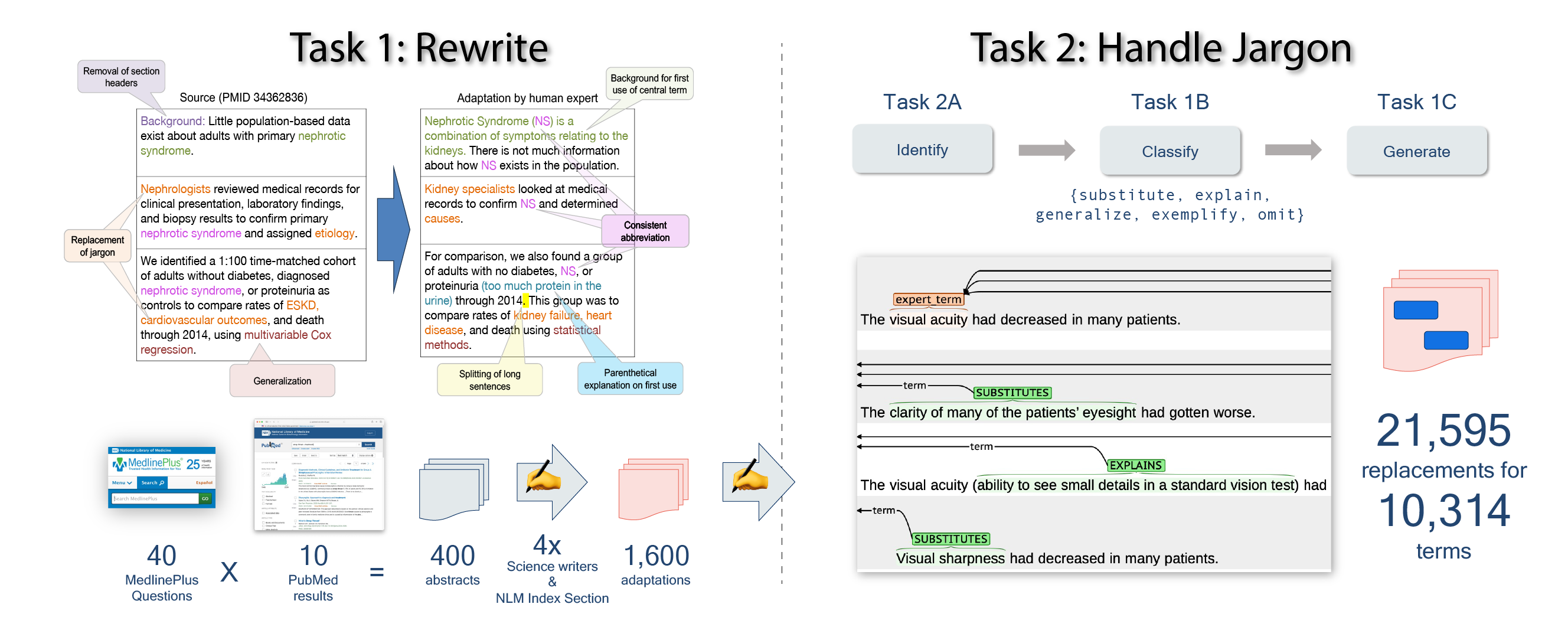}
\end{graphicalabstract}


\begin{keyword}



Evaluation\sep Large Language Models\sep Plain Language\sep Text Simplification
\end{keyword}

\end{frontmatter}


\section{Introduction}
The inability of patients to understand available information about their health significantly impacts outcomes~\citep{stableford2007plain,berkman2011low,king2010poor,kindig2004health}. While more information than ever is available on the web, the ``language barrier'' of professional-facing content, such as literature, can make it counterproductive~\citep{rosenberg2017online,white2009cyberchondria}. Though many consumer-facing knowledge bases exist, these are cumbersome and labor-intensive to update and thus typically do not include the latest medical knowledge from the literature. This is especially important for consumers seeking out the latest information about emerging health issues~\citep{seyyedhosseini2022comparing}.

The many recent successes of Artificial Intelligence suggest the possibility of automatically adapting text intended for a professional audience to a consumer audience.
Though Text Simplification has been widely studied in the open domain \citep{al2021automated}, the unique lexicon of biomedical literature makes it a distinct sublanguage~\citep{FRIEDMAN2002222}. In this respect, the task overlaps with Machine Translation \citep{stahlberg2020neural}. Recent examples of applying Deep Learning to Text Simplification specifically in the biomedical domain include \citet{van2019evaluating}, 
\citet{devaraj2021paragraph}, and \citet{flores2023medical}. However, the existing body of research evaluates systems with disparate datasets that are mined from imperfect sources, making progress difficult to measure.

With the aims of providing common benchmarks, encouraging progress, and rigorous expert evaluation, we hosted the Plain Language Adaptation of Biomedical Abstracts (PLABA) shared task at the Text REtrieval Conference (TREC), spanning 2023 and 2024.
This track included tasks for rewriting abstracts from the biomedical literature for the general public, as well as identifying expert terms and providing replacements or additional text for clarification.
By encouraging creation of systems that can coherently rewrite entire abstracts, but at the sentence level, we hope to allow the public to engage with the latest research, while providing fine-grained provenance of each statement to ensure transparency and reliability.

Across both years, twelve teams spanning twelve countries participated, submitting a total of 38 runs. Systems were diverse and powered by a variety of models, from Multilayer Perceptrons, to encoder-only transformer models such as BioBERT \citep{lee2020biobert} and RoBERTa \citep{liu2019roberta}, to encoder-decoder models like BART~\citep{lewis2020bart,yuan2022biobart} and T5 \citep{raffel2020exploring}, to modern, instruction-tuned decoder-only Large Language Models, including variants of Llama~\citep{touvron2023llama}, GPT~\citep{brown2020language,ouyang2022training}, Gemini~\citep{team2023gemini}, and Mistral~\citep{Mistral}. We find that the best performing systems can be highly factually accurate while making abstracts much more comprehensible to the general public. Still, occasional falsehoods and hallucinations mean that care must be taken in deploying such systems, especially given the risk of harm in the biomedical domain. Additionally, the task revealed there is work to be done in automatically evaluating both the simplicity and factuality of outputs. We hope the lessons from the PLABA shared task will inform future tasks and research directions.

\begin{center}
\begin{tabular}{p{12cm}}
\makecell[c]{\textbf{Statement of Significance}} \\ 
\toprule
\textbf{Problem}
\\ 
Patients and caretakers cannot keep up with 
biomedical research due to technical language and jargon.\\ \hline
\textbf{What is Already Known}
\\ 
Advanced Artificial Intelligence systems have sought to simplify professional text for a general audience.
\\ \hline
\textbf{What This Paper Adds}
\\ 
We stimulate methods research by providing a shared task and provide extensive manual review of outputs by biomedical experts to assess accuracy.
\\ \hline
\textbf{Who would benefit from the knowledge in this paper}
\\ 
Researchers using Artificial Intelligence to generate public-facing biomedical text output; clinicians who need to explain current research to patients. \\
\bottomrule
\end{tabular}
\end{center}

\section{Related Work}

Shared tasks on text simplification go back at least to Task 1 of the Sixth International Workshop on Semantic Evaluation (SemEval 2012) \citep{specia2012semeval}. This task challenged teams to rank replacement words by simplicity, given context for disambiguation. Lists of potential lexical replacements for this task came from SemEval 2007 Task 10, for which the goal was to provide substitutions, regardless of simplicity \citep{mccarthy2007semeval}. Both these tasks, however, were open domain, and did not focus on scientific text, which adds the complexity of technical language or jargon.

Beginning with a pilot in 2021, the SimpleText track~\citep{ermakova2021overview,ermakova2022overview,ermakova2023overview,ermakova2024overview} has been hosted yearly at the Conference and Labs of the Evaluation Forum (CLEF), a long-running European initiative with similar goals to TREC. As with PLABA, the SimpleText track aims to evaluate systems for simplifying scientific text. However, SimpleText has largely focused on Computer Science literature, with two relevant exceptions. In 2022, SimpleText data included a narrowly focused biomedical sub-corpus, comprising articles resulting from searching PubMed and Google Scholar for muscle health and hypertrophy. In 2024 a similar sub-corpus was included but expanding the search to articles on ``health and medicine.''

Each offering of SimpleText has had variations of three core tasks. Task 1 involves identifying passages of a scientific article that would be helpful for understanding by the general public \citep{sanjuan2022overview,sanjuan2023overview,sanjuan2024overview}.
Task 2 involves identifying difficult terms and providing definitions or explanations \citep{ermakova2022overview2,ermakova2023overview2,di2024overview}. This task is similar to PLABA Task 2, though PLABA's Task 2 additionally allows each term to be substituted, generalized, exemplified, or omitted, and further challenges teams to identify which of these are appropriate given the term and context. SimpleText Task 3 involves rewriting sentences or abstracts for a general audience, similar to PLABA's Task 1 \citep{ermakova2022overview3,ermakova2023overview3,ermakova2024overview3}.
The 2024 offering of SimpleText further included Task 4, which involves extraction of reported state-of-the-art results, though this only focused on Computer Science literature on the topic of Artificial Intelligence \citep{d2024overview}.

In addition to focusing on the biomedical domain, PLABA differentiates from SimpleText in the extent of manual evaluation. Evaluation of SimpleText Task 3 largely focuses on automatic, reference-based metrics, such as BLEU~\citep{papineni2002bleu} and SARI~\citep{xu2016optimizing}, and count-based readability measures, such as FKGL~\citep{kincaid1975derivation}. The 2024 offering additionally included analysis of spurious content (i.e. hallucinations) via alignment against sources. Extensive manual evaluation of a subset of outputs was performed for SimpleText 2022's Task 3, but only for the Computer Science corpus. Though manual evaluation was performed for SimpleText 2024's Task 3, reported results were limited to overall observations based on a small sample of outputs. PLABA's Task 1 adds to this work by providing in-depth manual evaluation for a purely biomedical test corpus, judging outputs for every sentence of 40 abstracts in 2023 and 400 abstracts in 2024.

Finally, the BioLaySumm track~\citep{goldsack2023overview,goldsack2024overview} has been hosted at the 22nd and 23rd Workshops on Biomedical Natural Language Processing (BioNLP 2023 and 2024), co-located with the 61st and 62nd Annual Meetings of the Association for Computational Linguistics (ACL 2023 and 2024), respectively. Like PLABA, BioLaySumm aims to evaluate and improve systems for conveying health information to consumers. However, the core BioLaySumm task is to summarize a complete scientific article into a short plain language paragraph, rather than to adapt each sentence of a scientific abstract. This assumes a large amount of information loss, giving readers only key takeaways. In contrast, by challenging teams to adapt each sentence of an abstract, PLABA involves preserving as much information as possible, in a manner that is more like `translating' from expert to plain language. This approach would offer readers direct provenance of each line of the plain language version, allowing them to check for consistency and learn more about the expert terminology.

\section{Tasks \& Data}

The PLABA track focused on public biomedical abstracts, sourced from PubMed.
Abstracts are useful because they are freely available and summarize the relevant findings from the latest literature, even if the complete articles are not open access. They are also small enough to be tractable and easily evaluated as whole units, but still large and varied enough to contain much of the types of text encountered in literature, such as biomedical terms, clinical jargon, and statistical details.
The inaugural offering of the PLABA track had a single task, which was complete, sentence-by-sentence rewriting of biomedical abstracts, which we refer to here as Task 1. For the second year, an additional task of identifying and replacing complex terms was added, which we refer to as Task 2.

\subsection{Task 1: Rewriting abstracts}

Task 1 challenged teams to rewrite biomedical abstracts at the sentence level and followed the format of the PLABA dataset~\citep{attal2023dataset}.

\subsubsection{Task Definition}

\begin{figure}[tb]
    \centering
    \begin{subfigure}[b]{0.48\columnwidth}
        \includegraphics[width=\linewidth]{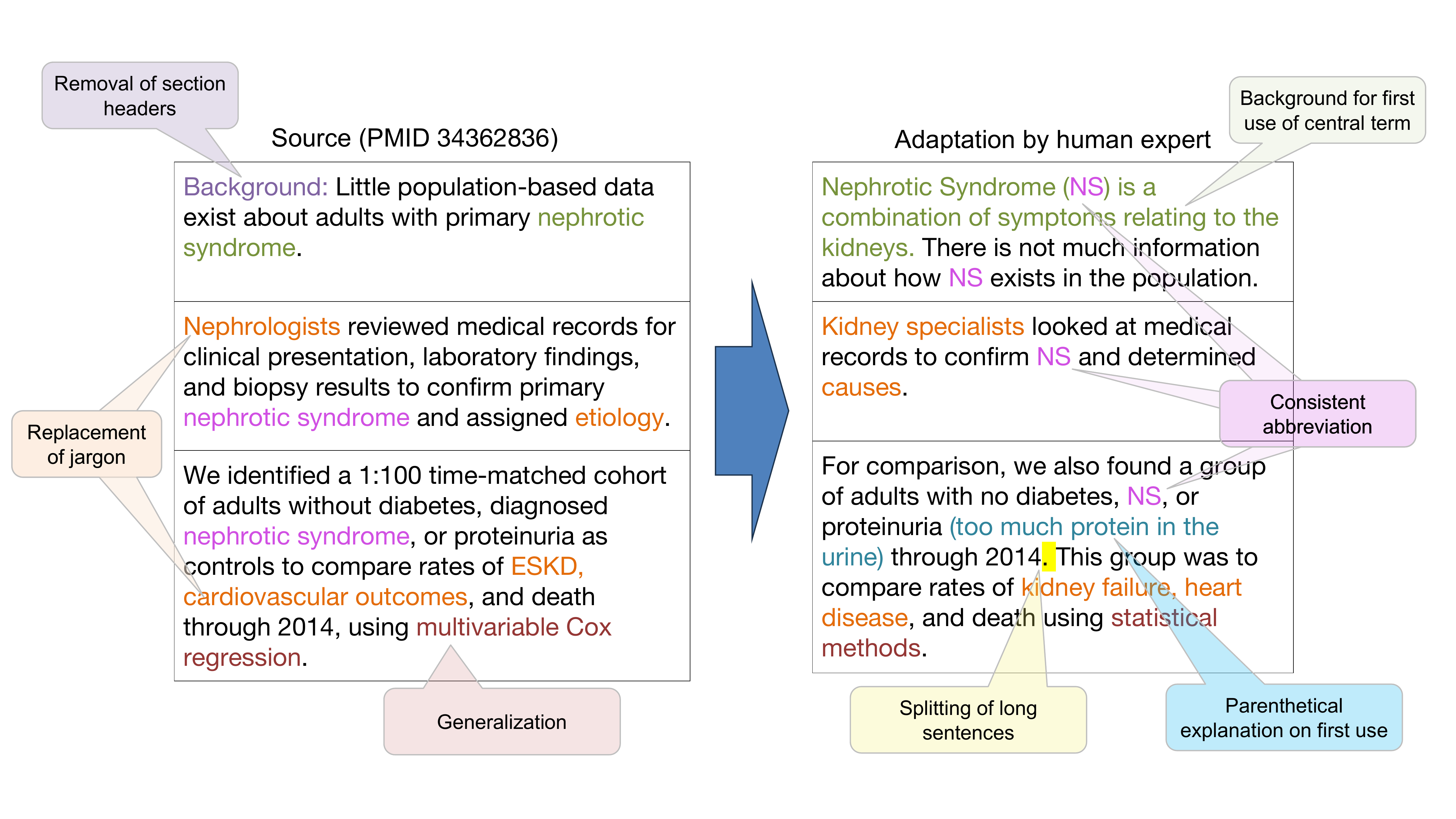}
        \caption{}
        \label{fig:task1}
    \end{subfigure}
    \begin{subfigure}[b]{.48\columnwidth}
        \includegraphics[width=\linewidth]{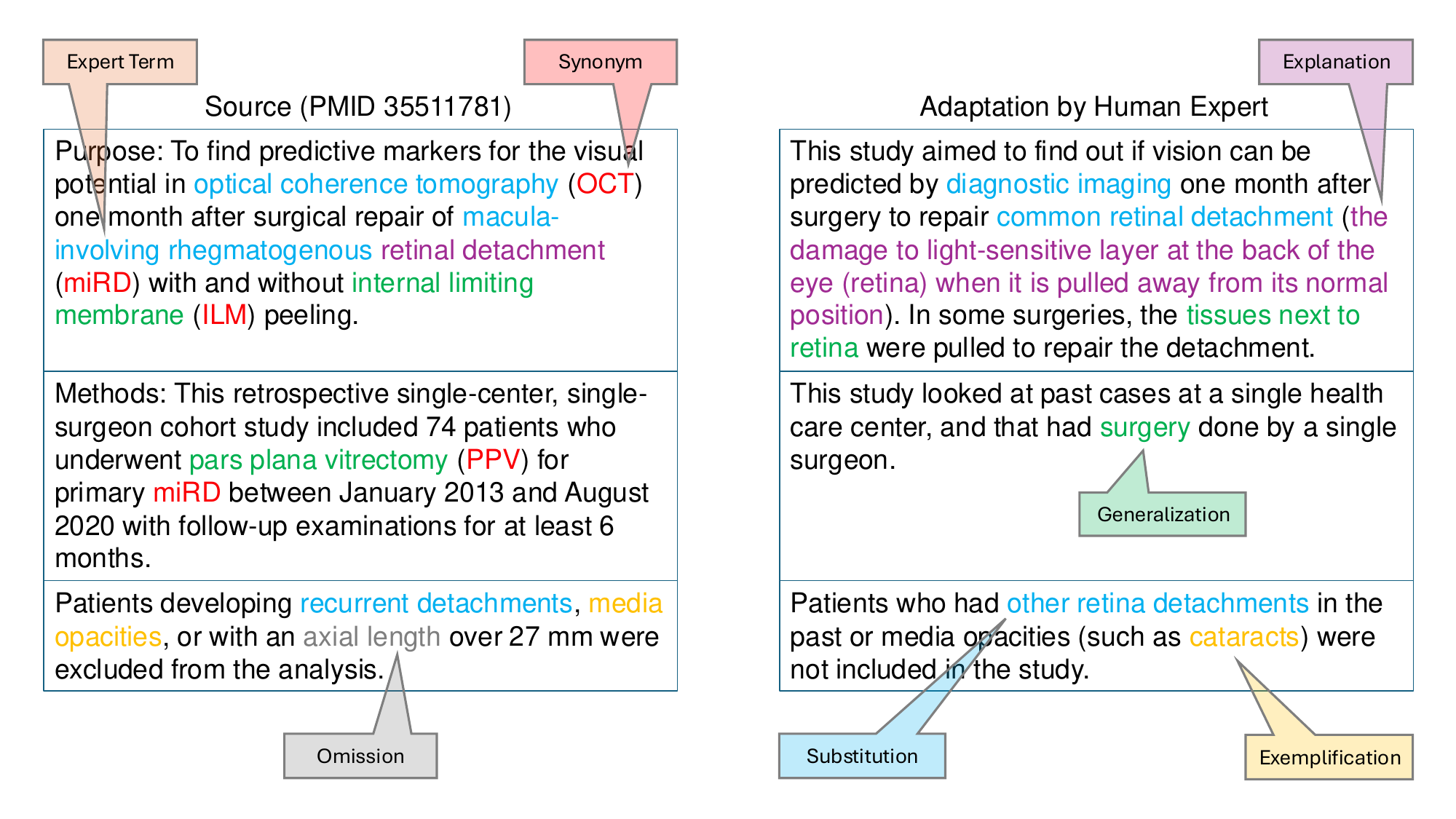}
        \caption{}
        \label{fig:task2}
    \end{subfigure}
    \begin{subfigure}[b]{.54\columnwidth}
        \includegraphics[width=\textwidth]{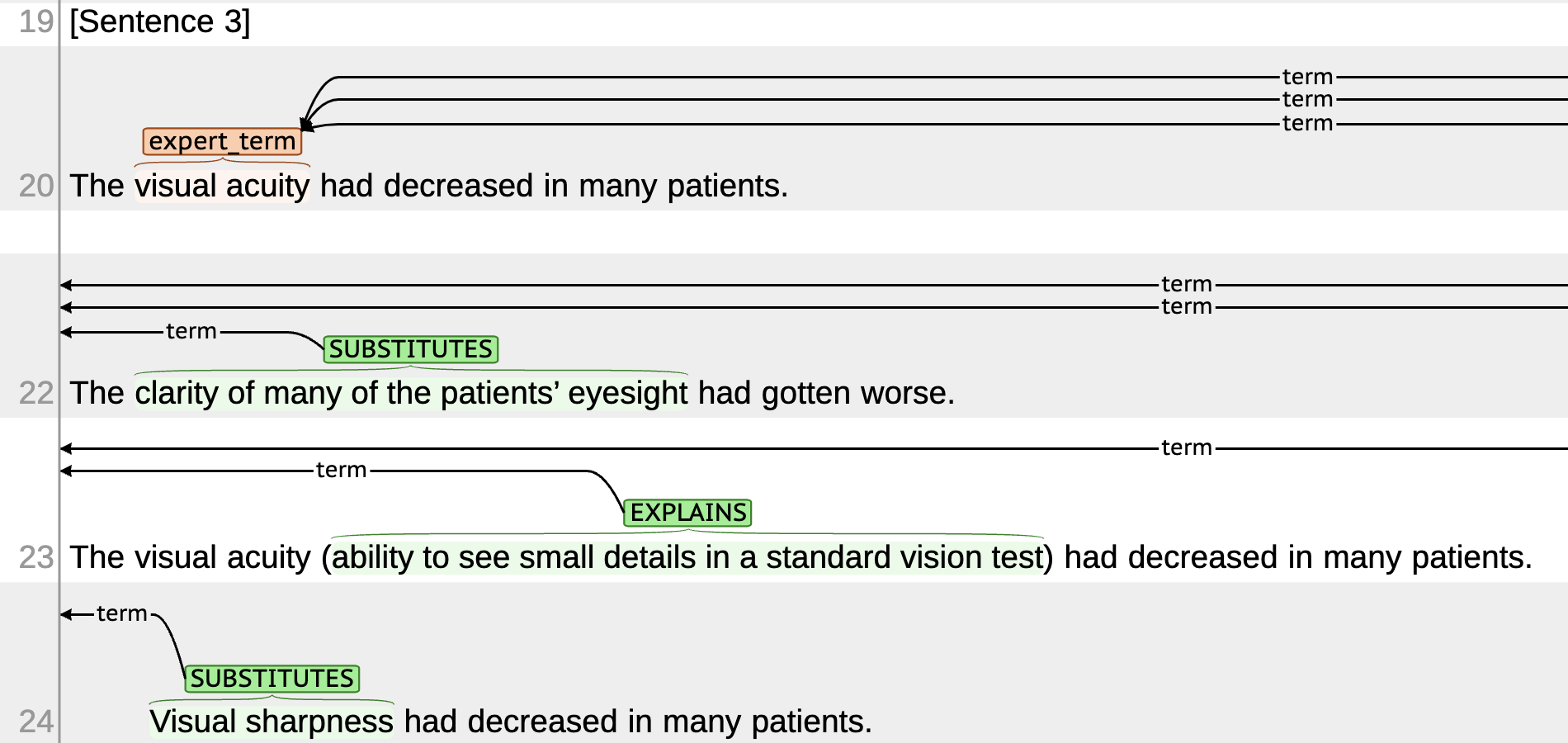}
        \caption{}
        \label{fig:ann}
    \end{subfigure}
    \begin{subfigure}[b]{.44\columnwidth} 
    \includegraphics[width=\textwidth]{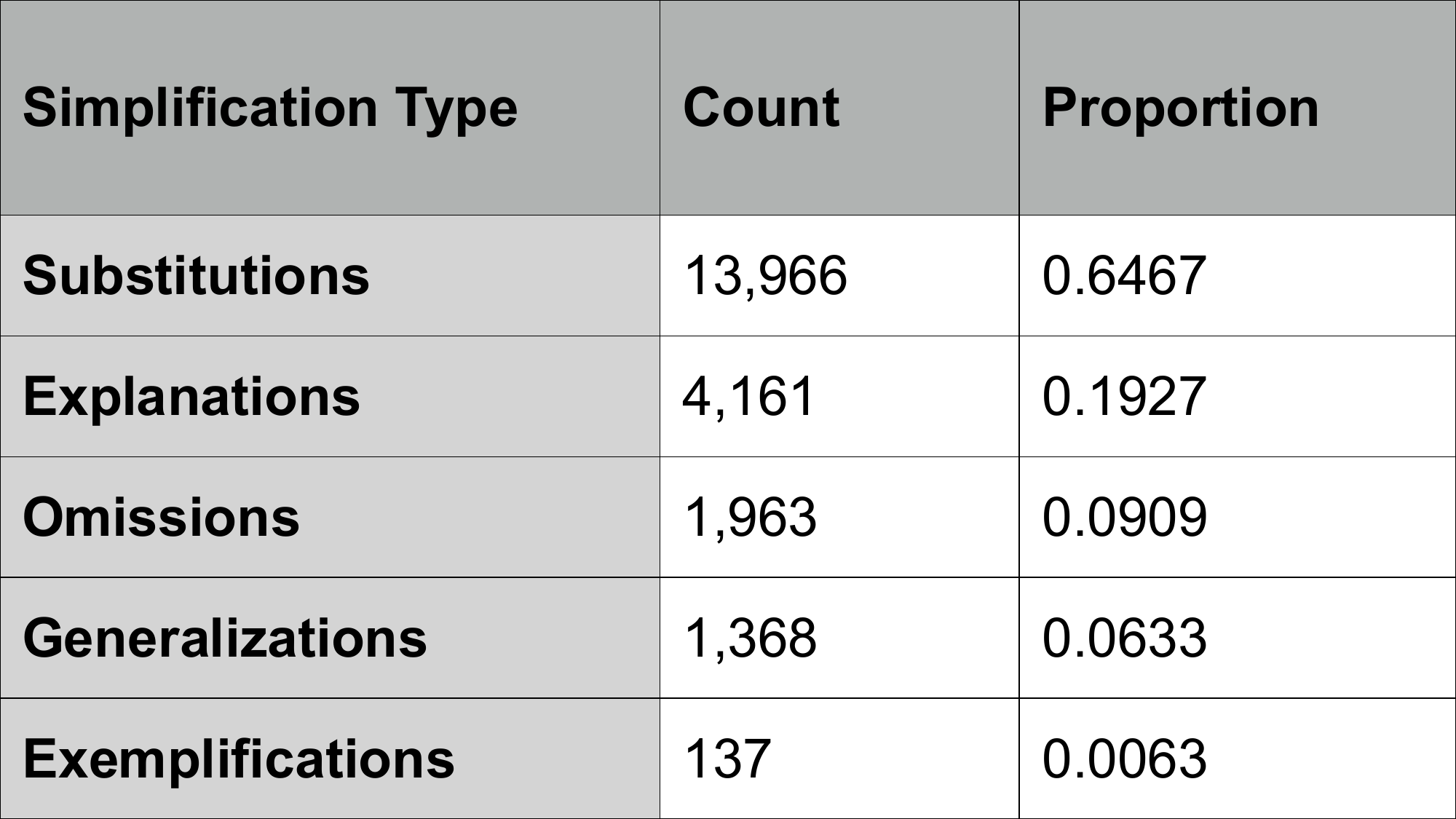}
    \caption{}
    \label{counts}
    \end{subfigure}
    \caption{PLABA tasks. In (a), an excerpt of a PLABA abstract and its manually written adaptation. Notable changes corresponding to annotation instructions are labeled. In (b), an PLABA abstract pair annotated for Task 2. Expert terms and their corresponding synonyms and simplifications are labeled. In (c), an example annotation of the Task 1 dataset, as seen on brat. Line 20 is the original sentence from an abstract; Lines 22 and 23 are from two adaptations. In each simplification, a replacement span has been identified, in one case being labeled as a substitution, and in the other being labeled as an explanation. In (d) counts and proportions of simplification types.}
    \label{fig:tsk}
\end{figure}

For a set of 40 consumer health questions, teams are presented with ten abstracts relevant to each question, each one split into sentences. The task is to create a sentence-aligned plain language adaptation of each abstract (Fig.~\ref{fig:tsk}). Being sentence-aligned means that output must be provided for each source sentence, but the entire rewritten abstract is also expected to read fluently as one document (e.g. not repeatedly explain expert terms). As in the PLABA dataset, sentences may be split (multiple output sentences for a single source sentence), and this is in fact encouraged when source sentences are long and complex. Sentences may also be dropped if they are deemed not relevant to a consumer's understanding of the abstract. However, sentences may not be merged (i.e. providing a single output sentence that spans the semantic content of multiple source sentences), to ensure there is a direct and consistent mapping from source to output for evaluation purposes.

\subsubsection{Training Data}

The training data for the task comprises the 750 abstracts initially published by ~\cite{attal2023dataset} chosen to answer 75 consumer health questions. Each abstract is manually adapted (rewritten) by at least one biomedical expert, for a total of 921 gold references. However, teams were welcome to use any other data at their disposal.

\subsubsection{Test Data}

As the test split from \cite{attal2023dataset} was already public and thus not suitable for shared task evaluation, in each year of Task 1 (2023 and 2024) we followed the same workflow as \cite{attal2023dataset} to choose an additional 40 questions and 10 abstracts for each question, totaling 400 abstracts. In the first offering (at TREC 2023), we created gold-standard manual reference adaptations of the 400-abstract test set. For robustness of automatic metrics, each abstract was adapted by four different annotators for the gold reference set. Three authors each adapted all 400 abstracts. The fourth adaptation for each abstract was performed by one of four contracted science writers. For the second offering (at TREC 2024), we did not create references for the test set and instead focused on manual evaluation (see \S\ref{sec:task1-2024}).

\subsection{Task 2: Identifying and Replacing Complex Terms}

For the second year of PLABA at TREC, we designed a new sub-task around identifying and replacing terms, rather than complete rewriting. One goal of this task was partly to lower the barrier to entry by including shorter generations and classification tasks, as the initial offering of Task 1 revealed that language models on the order of billions of parameters were necessary for competitive performance. Another goal was to generate finer-grained feedback for the lexical aspect of plain language adaptation.

\subsubsection{Task Definition}

An important step towards making biomedical text legible to consumers is replacing or explaining jargon or otherwise difficult terms. However, in existing text adapted for consumers, terms may be replaced or explained in various ways, depending on factors such as available synonyms, conceptual difficulty and importance of the term to the sentence or abstract. To code a set of replacement types, a group of three biomedical informatics experts examined the types of replacements in the PLABA dataset~\cite{attal2023dataset}. In multiple rounds of discussion, the experts finalized proposed types and suggested new types until a consensus was reached. This resulted in five types:

\newcommand{\classlabel}[1]{\texttt{\textsc{#1}}}

\begin{itemize}
    \item \classlabel{Substitute}: If the term is jargon for a more commonly understood concept, it can simply be replaced. For example, \textit{myocardial infarction} is the technical term for \textit{heart attack}. This may also apply to open-domain but more arcane words, for example \textit{indicate} can often be replaced with the synonym \textit{show}.
    \item \classlabel{Explain}: Other terms may not have any commonly understood equivalent, such as \textit{duodenum}. If such a term is important to the thought being conveyed, it can be left in the text but explained. Syntactically, this could take many forms, such as nonrestrictive clauses, parenthesis, or additional sentences. This type of replacement risks lengthening the original text, interrupt the flow of a sentence, or potentially introducing more complex jargon within the provided explanation.
    \item \classlabel{Generalize}: Many difficult terms have more broadly understandable hypernyms or superordinate concepts. For example, \textit{nucleic acid amplification test} may be generalized as \textit{lab test} if the specific type is not crucial to conveying the idea of the sentence.
    \item \classlabel{Exemplify}: A difficult concept may have widely known examples that can be provided alongside the term to elucidate it. For example, ``such as Parkinson's'' may be inserted after \textit{neurodegenerative diseases}. In a sense, this operation is the inverse of generalization. Care must be taken, however, that the examples provided do not break assumptions in the surrounding context.
    \item \classlabel{Omit}: If a difficult term is not necessary to understand the main thought that is being conveyed, the sentence may be rewritten in a way that avoids it entirely.
\end{itemize}

Note that \classlabel{Explain} and \classlabel{Exemplify} assume preservation of the original term in the new text, while \classlabel{Substitute}, \classlabel{Generalize}, and \classlabel{Omit} assume its removal.

Task 2 was broken into three sub-tasks. Teams could participate in any number of these, but each successive sub-task required participation in all previous sub-tasks.

\begin{itemize}
    \item \textbf{Task 2A - Identification:} Given the text of an entire abstract, systems should return a list of unique substrings representing words or phrases that a consumer would not understand.
    \item \textbf{Task 2B - Classification:} Given identified expert terms, classify how the terms should be replaced. As there may be multiple valid ways a human writer could handle replacing a given term, this is framed as a multi-label problem (that is, a binary classification problem for each of the five labels).
    \item \textbf{Task 2C - Generation:} Generate simplifications (replacement terms for \texttt{substitute} and \texttt{generalize} or additional text for \texttt{explain} and \texttt{exemplify}) given an abstract and the expert terms identified and classified within it. Note that if a term was classified only as \texttt{omit} in Task 2B, it would not have any generation for Task 2C.
\end{itemize}

\subsubsection{Data}

The Task 2 dataset is derived from the 400 abstracts and associated adaptations created for the 2023 Task 1 test data. Abstracts were aligned at the sentence level with their corresponding adaptations, then annotated by two authors using the brat rapid annotation tool\footnote{\url{http://brat.nlplab.org}} \citep{brat}, which involved selecting expert terms and linking them with their respective simplifications. Figure~\ref{fig:ann} shows an example of the brat interface during annotation. In total, the Task 2 dataset contains 10,314 expert terms (25.79 terms per abstract) and 21,595 simplifications. Table \ref{counts} displays counts of each simplification type. 
As many terms appear throughout an abstract in different forms (alternative wordings or abbreviations), annotators were also tasked with linking these synonyms. Annotations of replacements then only had to be performed for one representative term across the abstract. These annotations were then propagated to all synonyms during annotation postprocessing (see \ref{synonyms}).

The annotations exhibit a moderate inter-annotator agreement for both the identification task (0.5203 F1) and the classification task (0.4577 F1). Investigating disagreements for identification revealed that many related to minor differences in boundaries of annotated spans. We account for this during evaluation of Task 2A by requiring only 75\% overlap with a reference span to be considered correct. Upon investigating lower than ideal agreement for classification, we attribute it to multiple replacement types being valid for a given term in a given context, even though annotators had to choose one. We account for this in evaluation of Task 2B by framing it as a multilabel, rather than multiclass, classification problem. For further details on creation of the dataset for Task 2, see \cite{xia2025jebsfinegrainedbiomedicallexical}.


\section{Evaluation}

Here we describe how submissions were evaluated across all sub-tasks, including manual and automatic evaluation.

\subsection{Task 1 at TREC 2023}

For the initial offering of Task 1, we included both automatic and manual evaluation. As all evaluation was performed at the sentence level, we also developed a pipeline to automatically align submissions, allowing teams to make document-level submissions if they preferred.

\subsubsection{Automatic Evaluation}
\label{sec:task1-auto}

As the primary metric for automatic, reference-based evaluation, we adopt \texttt{SARI}, a metric specifically designed to assess simplification by including the source and balancing n-grams kept, inserted, and deleted in the references~\citep{xu2016optimizing}. 
Aside from the original implementation, several others exist, including in the EASSE package~\citep{alva2019easse} and in the Huggingface \texttt{evaluate} package~\citep{wolf2019huggingface}. 
Notably, the Huggingface implementation has several differences that purport to fix issues with the original implementation, leading to significantly different scores for some passages.\footnote{\url{https://huggingface.co/spaces/evaluate-metric/sari}} However, to our knowledge, only the original implementation has been shown to correlate with human judgments, and we thus use this implementation for the official results.\footnote{\url{https://github.com/cocoxu/simplification}}
For analysis of correlation of metrics with human judgments, we also compute  
\texttt{SAMSA}~\citep{sulem2018semantic}, a reference-free metric based on semantic structure, using the EASSE package~\citep{alva2019easse}, and \texttt{BLEU}~\citep{papineni2002bleu}, \texttt{ROUGE}~\citep{lin2004rouge}, and \texttt{BERTscore}~\citep{zhang2019bertscore} using the Huggingface \texttt{evaluate} package~\citep{wolf2019huggingface}.

\subsubsection{Manual Evaluation}

\begin{table*}[tb]
    \centering
    \small
    \begin{tabularx}{\textwidth}{p{0.05\linewidth}p{0.24\linewidth}p{0.6\linewidth}}
    \toprule
    \toprule
    \multicolumn{3}{c}{\textbf{Simplicity}} \\
\hline
SEN & Sentence simplicity & Are long, complex sentences appropriately split? \\
TRM & Term simplicity & Are expert terms in the source replaced with alternatives, or explained, either in this sentence or a previous sentence? \\
TAC & Term accuracy & Are substitutions and explanations of expert terms accurate? \\
FLU & Fluency & Does the output follow grammatical rules and read smoothly? \\
\textbf{SIM} & \textbf{Final simplicity} & \textbf{Average(SEN,TRM,TAC,FLU)} \\
\midrule
\midrule
    \multicolumn{3}{c}{\textbf{Accuracy}} \\
\hline
COM & Completeness & How much of the source information does the output provide? \\
FTH & Faithfulness & Do points made in the output match those of the source? \\
\textbf{ACC} & \textbf{Final accuracy} & \textbf{Average(COM,FTH)} \\
\midrule
\midrule
\textbf{FIN} & \textbf{Final score} & \textbf{Average(SIM,ACC)}\\
\bottomrule
\bottomrule
\end{tabularx}
\caption{
    Axes for manual evaluation of Task 1 at TREC 2023.}
    \label{tab:task1-axes1}
\end{table*}

For each of the 40 consumer questions, we randomly selected one of the ten abstracts retrieved for that question, creating a manual evaluation set of 40 abstracts representing all questions. Evaluation was divided into two main axes: \textit{simplicity} and \textit{accuracy} (Table~\ref{tab:task1-axes1}).  
Simplicity was further broken into four sub-axes: \textit{sentence simplicity}, \textit{term simplicity}, \textit{term accuracy}, and \textit{fluency}. Accuracy was broken into two sub-axes: \textit{completeness} and \textit{faithfulness}.
Note that we instructed annotators to judge faithfulness based on accurately carrying over statements made in the abstract, rather than comparing to general medical consensus. This was both because we aim to preserve the nuanced findings of various research studies and because medical consensus can be subjective.

Simplicity judgments were made for system output for each sentence of the set of 40 manual evaluation abstracts. We anticipated accuracy would be more labor-intensive due to the possibility of nuance and the need to research highly specialized topics discussed in the abstracts. We thus created a further restricted set of 3 sentences within each of these abstracts for accuracy judgments. To choose these sentences, each of three authors choose up to three lines of each abstract that they thought best answered or were most relevant to the consumer question for which the abstract was retrieved. Within each abstract, all lines were then ranked by the number of annotators choosing it, and the top three were chosen, with ties broken randomly.

We contracted four science writing experts to perform the annotations. Annotators were provided with rubrics for 3-point likert scale (-1, 0, 1) judgments for each axis and trained via a live video session. For final results, averages of likert scale values were linearly interpolated from (-1, 1) to (0, 100) for easier interpretation. As a pilot, all four annotated the first abstract, after which significant differences were discussed and questions clarified during a followup video session. During full evaluation of systems, a subset was double-annotated to compute inter-annotator agreement (see~\ref{apx:iaa}).

\subsubsection{Automatic Sentence Alignment}
\label{sec:aln}

Though PLABA is a sentence-level task, we anticipated that the importance of context would make document-level systems attractive. However, such systems typically do not assure sentence-alignment, which we require for evaluation. To give teams this option without the added burden of aligning output sentences, we provided an automatic alignment pipeline for document-level submissions. Additional details can be found in \ref{apx:aln}.

\subsection{Task 1 at TREC 2024}
\label{sec:task1-2024}

Though the first offering of Task 1 (at TREC 2023) included manually written-references to perform automated evaluation (\S\ref{sec:task1-auto}), analysis revealed poor correlation with manual evaluation, which we consider to be the gold standard (see \S\ref{sec:met}). For the second offering of Task 1 (at TREC 2024), we thus forewent expensive development of a reference set for the new test data, and instead diverted resources to further manual evaluation.

Additionally, based on results and feedback from contracted annotators from the first offering, we streamlined manual evaluation axes, resulting in four, rather than six, axes:

\begin{itemize}
    \item \texttt{SIM} (Simplicity): Is the output easy to understand for a non-expert?
    \item \texttt{ACC} (Accuracy): Does the output accurately reflect the source?
    \item \texttt{COM} (Completeness): Does the output minimize information loss?
    \item \texttt{BRV} (Brevity): Is the output as concise as possible?
    \item \texttt{FIN} \textbf{(Final score)}: Average of \texttt{SIM}, \texttt{ACC}, \texttt{COM}, and \texttt{BRV}.
\end{itemize}

This included merging ``sentence simplicity'' (SEN) and ``term simplicity'' (TRM) into an overall ``simplicity'' axis (SIM) and merging ``term accuracy'' (TAC) and ``faithfulness'' (FTH) into a simpler ``accuracy'' axis (ACC).

Further, for the second offering we deemed the ``fluency'' (FLU) axis unnecessary. Though fluency has traditionally been a basic aspect of manual evaluation of biomedical text simplification~\citep{ondov2022survey}, at the first offering of Task 1 at TREC 2023, 98\% of manually evaluated outputs were judged to have perfect fluency. In the intervening time between TREC 2023 and TREC 2024, we expected fluency of state-of-the-art models would only continue to improve thanks to scaling laws \citep{kaplan2020scaling}.

Instead of fluency, we introduce brevity (BRV), which captures whether systems produce output that conveys information as concisely as possible (taking into account how much information should be conveyed). This aspect of outputs had been conflated with SEN (sentence simplicity) in the initial offering, which caused confusion among annotators and contributed to poor inter-annotator agreement for this metric (see \ref{apx:iaa}). We thus deemed it better to provide it as a separate axis for the second offering.

Both the focus on manual evaluation (over manually created references for automatic evaluation) and streamlining of annotation axes and guidelines meant that annotators were able to evaluate all outputs of all systems, so it was not necessary to select a subset of abstracts for manual evaluation as at TREC 2023.

\subsection{Task 2 at TREC 2024}

We used automatic metrics for evaluating Tasks 2A (identification) and 2B (classification), and manual evaluation of generations from Task 2C.

\subsubsection{Automatic Evaluation}

Identification models were evaluated using F1 score against the union of terms identified by the two Task 2 data annotators. Classification models were evaluated according to union F1 score as well. Classification model scores were macro-averaged across the five simplification methods to account for the class imbalance in our data.

\subsubsection{Manual Evaluation of Task 2C}

Evaluations were performed manually along the same 4 axes as Task 1 at TREC 2024. Each axis was rated by contractors on a 3-point symmetric likert scale, which was interpolated to a 0-100 score for reporting.

\section{Results}

Both TREC 2023 and TREC 2024 were held as hybrid events at NIST's National Cybersecurity Center of Excellence facility in Rockville, MD, USA, in November of each year.
We will present the results in three parts: the two instances of Task 1 (at TREC 2023 and 2024), and the single instance of Task 2, at TREC 2024.

\subsection{Task 1 at TREC 2023}
\label{sec:task1-2023}
Four teams participated in the track. Initial submissions were due August 30\textsuperscript{th}, 2023, and results were announced October 18\textsuperscript{th}, 2023. Following the workshop, remaining resources allowed for an additional round of manual evaluation, for which teams were invited to submit an additional run if desired. For teams that did not submit an additional run, the run they had designated as second priority from the original submissions, if they had submitted more than one, was manually evaluated.

\subsubsection{Teams and Systems}

Team submissions and baselines are detailed in~\ref{sec:syst-task1-2023}. Each team was initially allowed three submissions, which they could rank in order of priority for manual evaluation. We also asked each team to provide short description of the strategy of each submission.



\subsubsection{Automatic Evaluation Results}

Results for automatic, reference-based evaluation using various metrics are shown in Table~\ref{tab:auto-task1-year1} (note we omit SAMSA due to lack of poor correlation with manual judgments, see \S\ref{sec:met}). Not surprisingly, fined-tuned systems did better in such metrics than zero-shot systems. However, this is not necessarily reflective of system quality, as there are many valid ways to adapt passages to plain language that may not be captured by the references, even with four-fold reference annotation.

\begin{table*}[h]
    \centering
    \small
    \begin{tabular}{lcccccc}
        \toprule
        \textbf{Submission} & \textbf{SARI} & \textbf{SARI-hf} & \textbf{BLEU} & \textbf{RL} & \textbf{R1} & \textbf{R2} \\
        \midrule
        PLABA\_1   & 35.90 & 39.00 & 28.95 & 46.75 & 53.25 & 28.63 \\
        PLABA\_2*   & 36.91 & 39.96 & 36.30 & 51.88 & 57.62 & 34.44 \\
        PLABA\_3*  & 23.31 & 23.31 & 5.30  & 9.21  & 11.07 & 4.81  \\
        PLABA\_4   & 44.12 & 49.11 & 59.51 & 69.59 & 71.87 & 55.92 \\
        PLABA\_5   & 42.65 & 49.58 & 61.93 & 71.23 & 73.39 & 58.20 \\
        \hline
        BeeManc\_1 & 40.58 & 42.97 & 42.05 & 56.45 & 60.20 & 39.75 \\
        BeeManc\_2 & 41.98 & 49.15 & 66.89 & 74.16 & 75.74 & 62.03 \\
        BeeManc\_3 & 42.12 & 49.92 & 68.19 & \textbf{75.33} & \textbf{76.91} & \textbf{63.15} \\
        BeeManc\_4 & 36.88 & 39.80 & 36.38 & 53.08 & 59.53 & 35.11 \\
        BoschAI\_1 & \textbf{44.69} & 49.03 & 58.50 & 68.98 & 71.81 & 54.98 \\
        BoschAI\_2 & 43.80 & 49.58 & 61.36 & 70.22 & 72.59 & 56.88 \\
        BoschAI\_3 & 32.84 & 35.94 & 19.97 & 36.44 & 44.79 & 20.51 \\
        MasonNLP\_1 & 39.47 & 42.92 & 42.44 & 56.86 & 60.88 & 38.98 \\
        MasonNLP\_2 & 39.97 & 43.47 & 43.49 & 57.16 & 60.95 & 41.90 \\
        MasonNLP\_3 & 38.38 & \textbf{50.89} & \textbf{68.34} & 75.00 & 76.78 & 63.12 \\
        PT3M\_1*   & 34.47 & 38.15 & 30.30 & 42.44 & 46.24 & 32.57 \\
        \bottomrule
        \multicolumn{7}{l}{\footnotesize{\makecell[l]{*Automatically aligned document-level submission (see \S\ref{sec:aln}).}}}
    \end{tabular}
    \caption{ Automatic evaluation results. SARI-hf is the HuggingFace implementation of SARI. RL, R1, and R2 refer to RougeL, Rouge1, and Rouge2, respectively.}
    \label{tab:auto-task1-year1}
\end{table*}

\subsubsection{Manual Evaluation Results}

\begin{table*}[h]
    \centering
    \small
    \setlength\tabcolsep{3pt}
    \renewcommand{\arraystretch}{0.9}
    \begin{tabularx}{\textwidth}{@{\extracolsep{\fill}} lccccccccc}
    \toprule
   & \multicolumn{5}{c}{\textbf{Simplicity}} & \multicolumn{3}{c}{\textbf{Accuracy}} \\
    \cmidrule(r){2-6} \cmidrule(r){7-9} 
\textbf{Submission} & SEN & TRM & TAC & FLU & \textbf{SIM} & COM & FTH & \textbf{ACC} & \textbf{FIN}\\
\hline
\multicolumn{9}{c}{\textbf{Round 1}} \\
\hline
PLABA\_1 & 91.45 & 86.84 & 91.22 & 93.53 & 90.76 & \textbf{95.73} & 94.02 & \textbf{94.87} & \textbf{92.82}\\
PLABA\_2* & \textbf{94.33} & 81.94 & 87.50 & \textbf{95.25} & 89.76 & 90.17 & 88.46 & 89.32 & 89.54\\
PLABA\_3* & 84.67 & 63.67 & 42.67 & 87.00 & 69.50 & 20.94 & 18.80 & 19.87 & 44.69\\
\hline
BeeManc\_1 & 92.84 & 82.33 & 64.20 & 91.57 & 82.74 & 79.49 & 70.51 & 75.00 & 78.87\\
BoschAI\_1 & 91.11 & 77.25 & \textbf{94.11} & 92.96 & 88.86 & 95.30 & \textbf{94.44} & \textbf{94.87} & 91.87\\
MasonNLP\_1 & 91.63 & \textbf{91.74} & 88.26 & 93.49 & \textbf{91.28} & 94.44 & 90.60 & 92.52 & 91.90\\
PT3M\_1* & 38.19 & 34.38 & 21.99 & 18.17 & 28.18 & 16.24 & 8.97 & 12.61 & 20.40\\
\hline
\multicolumn{9}{c}{\textbf{Round 2}} \\
\hline
PLABA\_1\textsuperscript{\textdagger} & 83.14 & 87.30 & 96.88 & \textbf{98.73} & 91.51 & 98.72 & 96.58 & 97.65 & 94.58\\
PLABA\_4 & 82.71 & 84.35 & 93.93 & 95.68 & 89.17 & 95.73 & 96.15 & 95.94 & 92.55 \\
PLABA\_5 & 78.67 & 76.92 & 90.33 & 97.09 & 85.75 & 97.01 & 96.58 & 96.80 & 91.27 \\
\hline
BeeManc\_4 & 79.91 & 74.02 & 92.49 & 97.58 & 86.00 & 96.58 & 95.30 & 95.94 & 90.97\\
BoschAI\_2 & 79.49 & 76.81 & 90.33 & 98.83 & 86.36 & 95.30 & 97.01 & 96.15 & 91.26\\
MasonNLP\_2 & 79.78 & 85.17 & 89.59 & 92.34 & 86.72 & 93.16 & 86.75 & 89.96 & 88.34\\
\hline
Manual\_1 & \textbf{96.67} & \textbf{98.02} & 96.67 & 97.65 & \textbf{97.25} & 92.74 & 93.16 & 92.95 & 95.10\\
Manual\_2 & 94.35 & 96.12 & \textbf{98.71} & 97.53 & 96.68 & 96.58 & 97.86 & 97.22 & \textbf{96.95}\\
Manual\_3 & 84.78 & 87.59 & 92.86 & 97.54 & 90.69 & 96.15 & 94.87 & 95.51 & 93.10\\
Manual\_4 & 81.79 & 86.08 & 94.90 & 95.71 & 89.62 & 99.15 & \textbf{99.15} & \textbf{99.15} & 94.39\\
\hline
Manual\_avg & 89.40 & 91.95 & 95.79 & 97.11 & 93.56 & 96.16 & 96.26 & 96.21 & 94.88\\
\bottomrule
\multicolumn{9}{l}{\makecell[l]{\footnotesize{*Automatically aligned document-level submission (see \S\ref{sec:aln}).} \\
\footnotesize{\textsuperscript{\textdagger}Reevaluated by second round evaluators for comparison with round one.}}}
\end{tabularx}
\caption{
    Results of the first and second rounds of manual evaluation of Task 1 at TREC 2023.}
    \label{tab:man1}
\end{table*}

\begin{figure}[htb]
    \centering
    \begin{subfigure}[b]{.48\textwidth}
        \includegraphics[width=\textwidth]{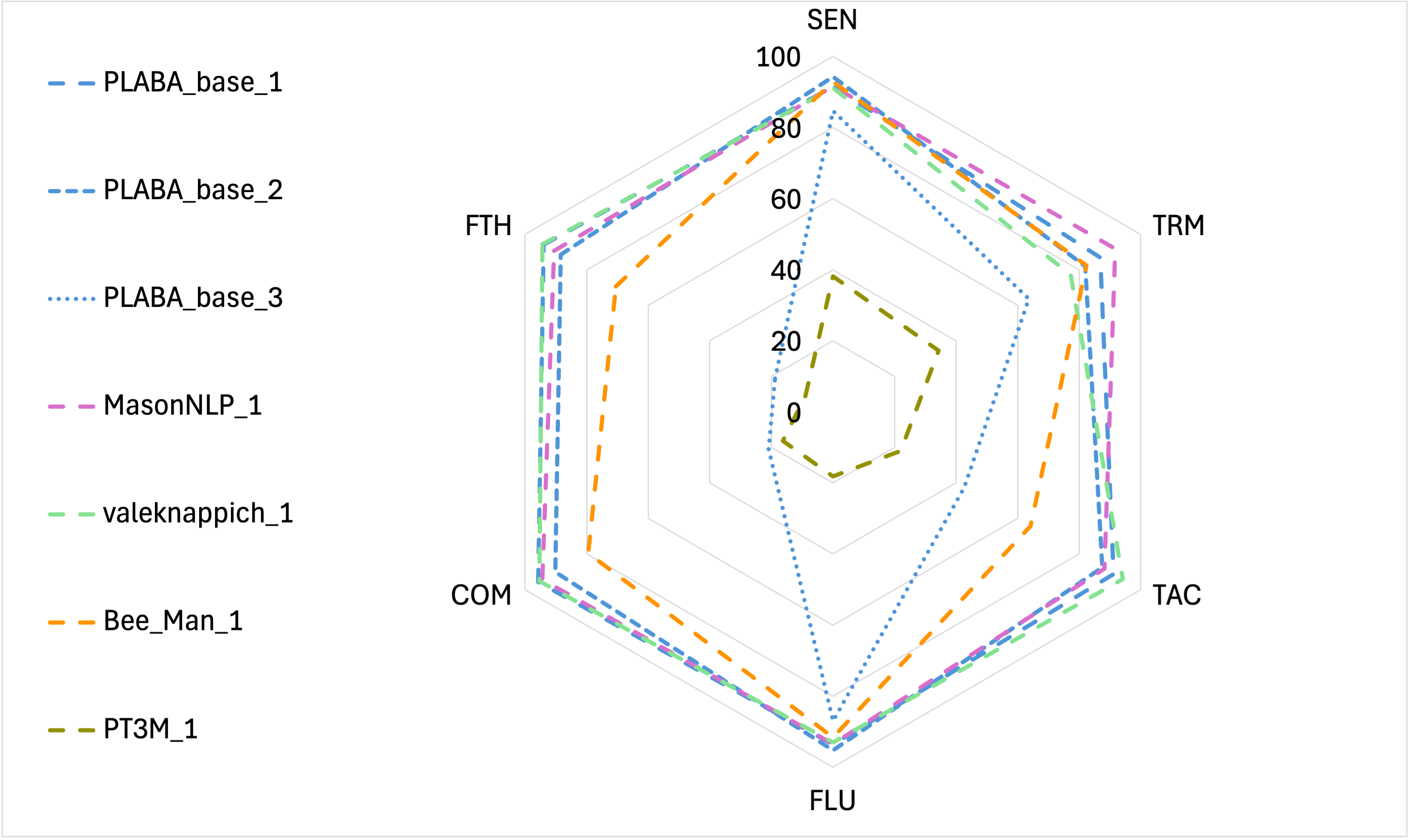}
        \caption{}
        \label{fig:task1-2023-man}
    \end{subfigure}
    \begin{subfigure}[b]{.48\textwidth}
        \includegraphics[width=\textwidth]{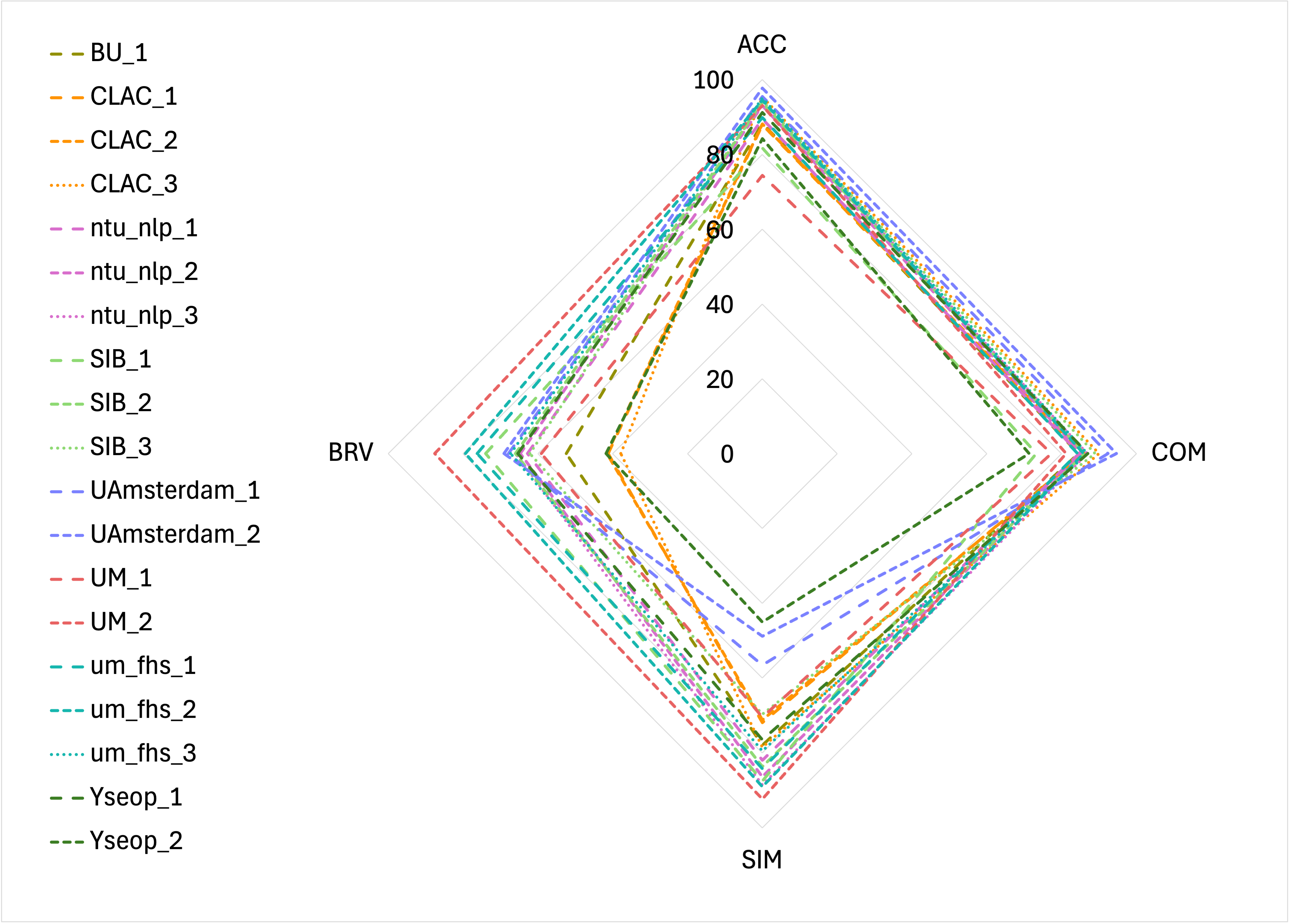}
        \caption{}
        \label{fig:task1-2024-man}
    \end{subfigure}
    \caption{Radar charts of manual evaluation results for Task 1 at (a) TREC 2023 and (b) TREC 2024.}
\end{figure}

As further resources were available following the first round of manual evaluation, we invited teams to submit an additional run for a second round of manual evaluation. As only one team (BeeManc) did so, we manually evaluated the second priority runs for the other teams, except PT3M, who had only submitted one run. Resources also allowed us to include two additional baselines, both fine-tuned (\texttt{PLABA\_4} and \texttt{PLABA\_5}), and to include the manually written reference adaptations for manual evaluation, creating a human performance baseline. Finally, since different annotators were used versus the first round, we included the top-performing system (\texttt{PLABA\_1}) for re-evaluation as a point of reference.
Note, however, that these results are not directly comparable to those in round 1 because of the change in annotators and timing (after the conference rather than before).
Results of both rounds of manual evaluation are shown in Table~\ref{tab:man1} and visualized in Figure~\ref{fig:task1-2023-man}.
During the second round of evaluation, resources allowed for giving annotators overlapping assignments for three abstracts in order to estimate inter-annotator agreement (see~\ref{apx:iaa}). 

\subsubsection{Analysis of Metrics}
\label{sec:met}

\begin{figure}
\centering
\begin{subfigure}[b]{0.8\textwidth}
    \includegraphics[width=\textwidth]{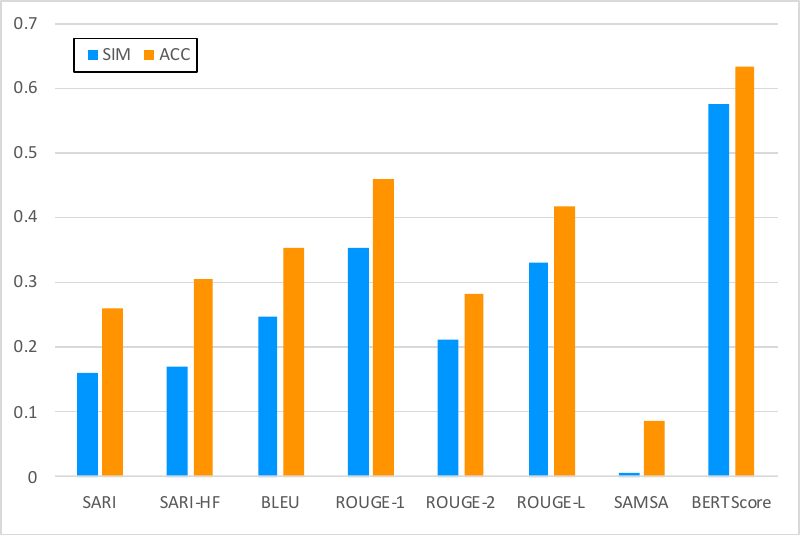}
    \caption{}
    \label{fig:cor}
    \end{subfigure}
    \begin{subfigure}[b]{0.48\textwidth}
        \includegraphics[width=1\linewidth]{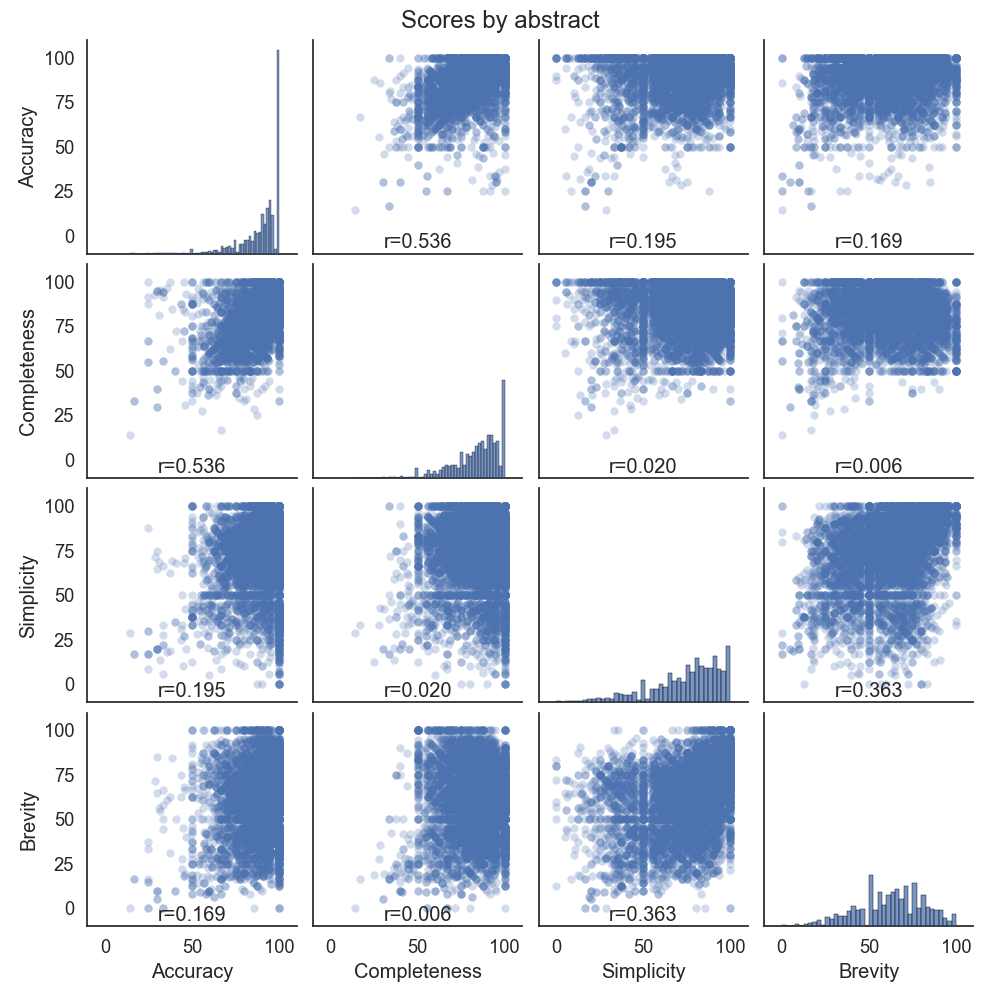}
        \caption{}
        \label{fig:corr-task1}
    \end{subfigure}
    \begin{subfigure}[b]{0.48\textwidth}
        \includegraphics[width=1\linewidth]{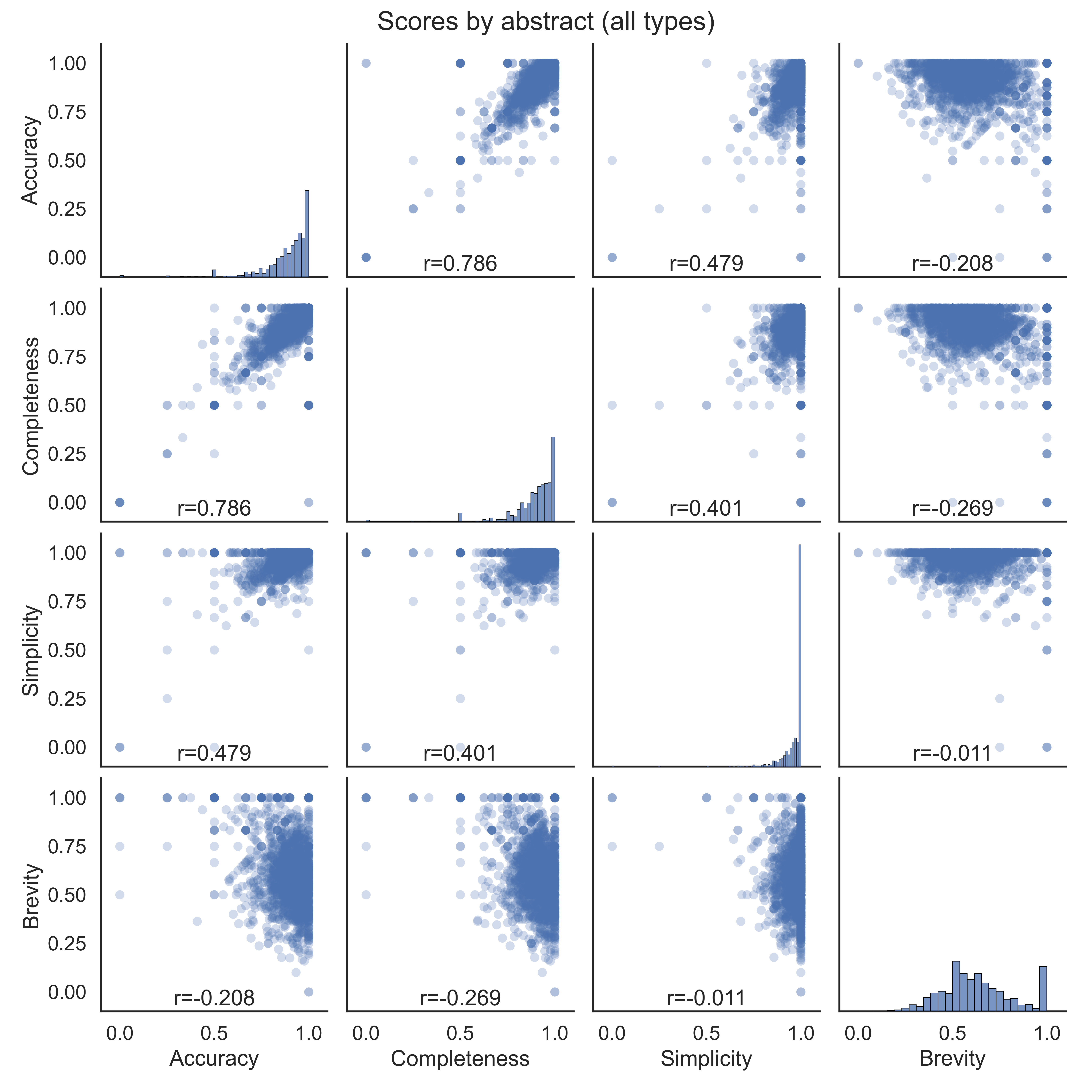}
        \caption{}
        \label{fig:corr-task2}
    \end{subfigure}
    \caption{In (a), correlation of automatic metrics with manually evaluated Simplicity (SIM) and Accuracy (ACC). For evaluating relationships between manual evaluation axes, pairwise scatterplots of the four manual evaluation axes are shown for (b) Task 1 and (c) Task 2, both at TREC 2024. Each point represents output for one abstract by one system, with values averages across sentences (Task 1) or terms (Task 2). Each scatterplot is labeled with its Pearson correlation value (r). Histograms on the diagonal show the distributions of scores for each of the axes.}
\end{figure}

To assess the utility of automated metrics, we compute their correlation with manual judgments. Correlations of all computed metrics are shown in Figure~\ref{fig:cor}. Scatterplots with regression lines for each metric are shown in \ref{apx:met}.

\subsection{Task 1 at TREC 2024}

\begin{table}[tb]
    \centering
    \small
    \begin{tabular}{lccccc}
        \hline
        \textbf{Submission} & ACC & COM & SIM & BRV & \textbf{FIN} \\
        \hline
        BU\_1 & 88.31 & 84.47 & 77.92 & 52.40 & 75.78 \\
        CLAC\_1 & 87.88 & 85.67 & 71.88 & 41.22 & 71.66 \\
        CLAC\_2 & 88.18 & 86.03 & 71.18 & 41.55 & 71.74 \\
        CLAC\_3 & 95.37 & 89.96 & 78.50 & 37.82 & 75.41 \\
        ntu\_nlp\_1 & 89.48 & 84.63 & 81.97 & 62.72 & 79.70 \\
        ntu\_nlp\_2 & 93.07 & 85.37 & 86.42 & 64.68 & 82.38 \\
        ntu\_nlp\_3 & 93.95 & 86.81 & 88.60 & 65.86 & 83.81 \\
        SIB\_1 & 81.70 & 73.24 & 87.56 & 73.92 & 79.10 \\
        SIB\_2 & 93.74 & 86.14 & 83.63 & 65.94 & 82.36 \\
        SIB\_3 & 94.33 & 88.91 & 69.65 & 62.10 & 78.75 \\
        UAmsterdam\_1 & 95.46 & 92.38 & 56.53 & 67.81 & 78.04 \\
        UAmsterdam\_2 & \textbf{97.72} & \textbf{94.66} & 48.81 & 69.06 & 77.56 \\
        UM\_1 & 74.40 & 77.25 & 70.44 & 59.03 & 70.28 \\
        UM\_2 & 93.07 & 81.18 & \textbf{92.32} & \textbf{87.55} & \textbf{88.53} \\
        um\_fhs\_1 & 89.85 & 84.27 & 84.21 & 76.18 & 83.63 \\
        um\_fhs\_2 & 94.47 & 85.88 & 89.09 & 79.30 & 87.18 \\
        um\_fhs\_3 & 95.04 & 86.81 & 79.49 & 67.28 & 82.15 \\
        Yseop\_1 & 91.17 & 87.00 & 76.54 & 65.33 & 80.01 \\
        Yseop\_2 & 84.24 & 71.32 & 45.00 & 41.77 & 60.58 \\
        \hline
    \end{tabular}
    \caption{Results of manual evaluation of Task 1 at TREC 2024.}
    \label{tab:task1-2024-man}
\end{table}

Submissions to Task 1 at TREC 2024 were due September 20\textsuperscript{th}, 2024 and preliminary results were announced November 10\textsuperscript{th}, 2024. With the increase in interest from the previous year, we expected more submissions. We thus did not develop any baselines for the second offering of this task and instead focused on developing baselines for the newly introduced Task 2 (see \S\ref{sec:task2}). Eight teams participated in the second offering of Task 1, with 19 total submissions. Results of manual evaluation of all submitted systems, based on sentence-level outputs for all 400 test abstracts, are shown in Table~\ref{tab:task1-2024-man}, and visualized in Figure~\ref{fig:task1-2024-man}. 

Additionally, we investigate potential tradeoffs in system strengths by computing correlations among each pair of manual judgment axes (Fig.~\ref{fig:corr-task1}). Though individual judgments were made at the sentence level, this would be too fine-grained for correlation analysis since judgments are discrete (-1, 0, or 1). System-level aggregation, however, would leave too few points to see evidence of patterns. We thus aggregate scores at the abstract level for this analysis. As can be seen from the scatterplots and correlation values, accuracy is strongly related to completeness, but other pairs are not highly correlated. Interestingly, the lowest r value is for completeness versus brevity, which one might expect to be in tension. However, even this value is still slightly positive.

\subsection{Task 2 at TREC 2024}
\label{sec:task2}

Six teams participated in Task 2 at TREC 2024, with nine total submissions. Submissions were due September 20\textsuperscript{th}, and preliminary results were announced November 10\textsuperscript{th}, 2024.
Team submissions and baseline systems are detailed in \ref{sec:syst-task2-2024}. Each team was allowed
three submissions, which they could rank in order of priority for manual
evaluation. Tasks 2B and 2C were optional, though participation in 2C required participation in 2B. We also asked each team to provide a short description of each submission, as well as any base LLMs and training data used.

\subsubsection{Evaluation Results}

All results for Task 2 at TREC 2024, including automatic metrics for Tasks 2A and 2B and manual judgments for Task 2C, are shown in Table \ref{tab:man2}. As for Task 1, we compute correlations among each pair of manual evaluation axes for Task 2C (Fig.~\ref{fig:corr-task2}). Again, accuracy is strongly related to completeness; however, in this case, brevity and completeness exhibit a negative correlation, illustrating the tradeoff between outputs being complete and brief. Pairwise scatterplots and correlations broken down by replacement type are shown in \ref{sec:corr-task2-types}.

\begin{table}[]
\small
\begin{tabular}{lccccccc}
\toprule
\textbf{}   & \textbf{Task 2A} & \textbf{Task 2B} & \multicolumn{5}{c}{\textbf{Task 2C (manual eval., 0-100)}}                    \\ \cmidrule{4-8} 
\textbf{Submission}            & \textbf{(F1)}    & \textbf{(F1)}    & ACC & COM & SIM & BRV & \textbf{FIN} \\
\hline
PLABA\_2A\_1 & 0.2487 & - & - & - & - & - & - \\
PLABA\_2A\_2 & \textbf{0.5255} & - & - & - & - & - & - \\
PLABA\_2A\_3 & 0.4085 & - & - & - & - & - & - \\
PLABA\_2B\_1 & 0.3399 & 0.3413 & - & - & - & - & - \\
PLABA\_2B\_2 & 0.4009 & 0.3363 & - & - & - & - & - \\
\hline
BU\_1       & 0.0459           & 0.2868           & 80.56         & 82.79         & 95.40         & \textbf{87.43}         & 86.55         \\
CLAC\_1     & 0.4410            & 0.3317           & 93.39         & 93.68         & 97.75         & 64.26         & \textbf{87.27}         \\
CLAC\_2     & 0.3767           & 0.1865           & \textbf{95.57}         & \textbf{95.17}         & 98.54         & 58.98         & 87.06         \\
IITH\_1     & 0.1956           & 0.2759           & -             & -             & -             & -             & -             \\
UM\_1       & 0.4787           & 0.3180           & -             & -             & -             & -             & -             \\
ntu\_nlp\_1 & 0.4885           & \textbf{0.3931}           & 86.59         & 86.77         & 95.13         & 57.29         & 81.44         \\
ntu\_nlp\_2 & 0.4431           & 0.3715           & 86.48         & 87.84         & 94.22         & 53.25         & 80.45         \\
ntu\_nlp\_3 & 0.4518           & 0.3287           & -             & -             & -             & -             & -             \\
Yseop\_1    & 0.5036           & 0.1854           & -             & -             & -             & -             & -      \\
\bottomrule
\end{tabular}
    \caption{
    All results for Task 2 at TREC 2024, including automatic metrics for Tasks 2A and 2B and manual judgments for Task 2C. Missing values mean the submission did not participate in that sub-task. Task 2B F1 scores are macro-averaged over the five classes.}
    \label{tab:man2}
\end{table}

\section{Discussion}

Here we discuss key lessons that we hope will inform future iterations of this task and potentially other tasks in the Large Language Model era.

\subsection{Bridging the Gap from Sentences to Documents}

Though prior shared tasks have involved either sentence-level or document level plain language generation, the PLABA track is, to our knowledge, the first shared task that evaluated biomedical text simplifications at the sentence level while accounting for the context of the entire abstract. This posed unique challenges for test data generation, system submission, and evaluation.
Document-level systems generally do not ensure sentence-aligned output for rewriting tasks, which can complicate evaluation of fine-grained details. In this shared task, automatically aligning system output to sentences was a reasonable compromise, with state-of-the-art alignment tools built for Machine Translation proving valuable (though in practice only one of the ten submissions, and two of the three baselines, utilized this option). Further, our autoregressive \texttt{PLABA\_1} baseline showcased how Large Language Models can be strategically prompted to include context while adapting individual sentences.

Still, challenges remain for both evaluation and generation. Due to its reliance on dynamic programming, our automatic alignment pipeline cannot account for transposition of sentences, which may be useful when rewriting a document for a lay audience. By choice, we also did not allow source sentences to be merged, as it is not yet clear how to standardize sentence-level evaluation with this possibility. Yet, merging sentences is also a potentially useful operation. Future work can explore evaluation methods that capture how well fine-grained semantic content is preserved across document-level adaptations.

Finally, though dropping sentences was allowed for this task, we found manual writers used it very rarely, and sentence-level submissions from teams did not use it at all. For document-level submissions that contain dropped sentences after being automatically aligned, we did not have an explicit evaluation metric for this operation. Future iterations of the task could either eliminate this possibility (requiring output for every sentence) or treat sentence dropping as a binary classification sub-task to capture performance. 

\subsection{Insufficiency of Automated Metrics}

The main PLABA task is similar to the longstanding tasks of Machine Translation and Text Simplification. Machine Translation, however is fairly restricted semantically, whereas Plain Language Adaptation has much more freedom to rephrase, add content, and remove content. PLABA also deviates from more traditional forms of Text Simplification, which until recently have largely revolved around (1) lexical substitutions and (2) atomic operations on syntax, both of which preserve much of the dependency structure of a given sentence~\citep{ondov2022survey}. Whether automatic, reference-based metrics that have worked for the latter tasks would extend to PLABA was thus an open question. We find that n-gram based metrics generally correlated poorly with manual judgments of both simplicity and accuracy of content, with metrics specifically designed for the task of Text Simplification (\texttt{SARI} and \texttt{SAMSA}) notably having the worst correlations. This is despite including four unique, manually written reference adaptations for each source sentence. \texttt{BERTScore} had much higher correlation with both simplicity and accuracy of content, concurring with findings of \cite{alva2021suitability}.

Still, the discrepancies between system rankings by automatic metrics versus manual judgments highlight that there are many possible ways to rewrite sentences for this task, and these may not captured even after considerable effort to write several high-quality versions. Further, small changes in similarity to a reference could have outsize influence on the message a healthcare consumer takes from an adaptation, and further work is needed to assess how well any automated metric captures possible harms. Future work can also focus on developing new ways to automatically asses how easy to read output is and how well it captures the factual content of the original, without relying on word or n-gram similarity.

\subsection{Factuality and Hallucinations}

A consequence of using pretrained language models for transfer learning and zero-shot applications is the mismatch between their explicit training objective (maximizing the likelihood of the next word given a corpus) and more specialized downstream tasks, which often implicitly require knowledge or reasoning. This leads to the well-known problem of ``hallucination,'' or the output of cogent but unfounded text. In biomedical text simplification, the shift from the rule-based era to the neural era represented a marked shift in error profiles, from chiefly errors of grammatically (despite factual accuracy) to chiefly errors of factual accuracy (despite often perfect fluency)~\citep{ondov2022survey}. This phenomenon has been exacerbated by LLMs, which can now fabricate entire abstracts, complete with internally coherent study details and imagined citations.

In this shared task, however, we find that, generally, the most fluent systems are also rated to be highly factual. In fact, when blindly manually evaluated alongside reference adaptations (which were manually written by biomedical experts) for Task 1 at TREC 2023, the top-performing \texttt{PLABA\_1} system even exceed the average manual score for completeness (COM), faithfulness (FTH), and their combination, accuracy (ACC) (see \S\ref{sec:task1-2023}). Top systems from Task 1 at TREC 2024 pushed accuracy measures even higher (though we caution against direct comparisons because of the change in annotators across years). This suggests that state-of-the-art language models, guided by the context of original abstracts, can produce very accurate information on detailed biomedical topics.

Still, this shared task also revealed that, even when generation is narrowly focused on one source line, large, convincing hallucinations can still occur (as described in~\ref{apx:hal}). These may be all the more insidious if users learn to trust systems that are largely accurate. Health information provided to consumers can be highly actionable, and there is thus a high potential for harm even in rare edge cases. We must thus be vigilant in evaluating and deploying systems that provide such information.
Future work should investigate more rigorous ways to detect and mitigate hallucinations and automatically assess factuality to ensure these kinds of errors do not go undetected.

\section{Conclusion}

Two years of the Plain Language Adaptation of Biomedical Abstracts (PLABA) track challenged teams to rewrite biomedical abstracts for the general public, and to identify and replace expert terms in appropriate ways. The track drew a diverse group of teams from around the world, with some attending the conference in person and some taking advantage of the virtual option. Submissions showcased a wide variety of systems, with language models of many types and sizes, different prompting strategies, and custom training pipelines. The track also featured baseline systems running the gamut from ruled-based systems for expert term identification to state-of-the-art instruction-following pretrained transformers for end-to-end rewriting of abstracts. Though systems had a wide range of results, especially in manual evaluation, the best-performing systems neared or surpassed human levels of factual accuracy while performing near-human levels of simplification. These systems in their current form may already be able to help consumers interpret the latest biomedical research for better healthcare engagement and literacy.
Accurate consumer-oriented biomedical texts with each line attributable to a trustworthy, peer-reviewed abstract could also provide a valuable intermediate resource for abstractive consumer question answering or Retrieval Augmented Generation~\citep{lewis2020retrieval}.
Still, we urge caution when deploying such systems, as even the best systems from this track still made errors, and even minor errors may have the potential for harm in the biomedical setting. Erring on the side of less simple, but more accurate systems, using our evaluations as a guide, will allow researchers to follow a ``progressive caution'' approach~\citep{goodman2021ethics}. We hope the lessons from this task and the resulting systems will inform and inspire future work in the domain of consumer-focused biomedical text generation.

\section{Acknowledgements}

This work was supported in part by the intramural research program of the National Library of Medicine, National Institutes of Health, Bethesda, MD, USA, and utilized the computational resources of the NIH HPC Biowulf cluster (\url{https://hpc.nih.gov}).


  \bibliographystyle{elsarticle-harv} 
  \bibliography{references}






\appendix

\renewcommand{\thesection}{Appendix~\Alph{section}}

\setcounter{table}{0}
\setcounter{figure}{0}
\renewcommand{\thealgorithm}{A.\arabic{algorithm}}

\section{Assignment of System Sentences From Alignments for Document-level Submissions}
\label{apx:aln}

\begin{figure}[tb]
\centering
\includegraphics[width=0.9\textwidth]{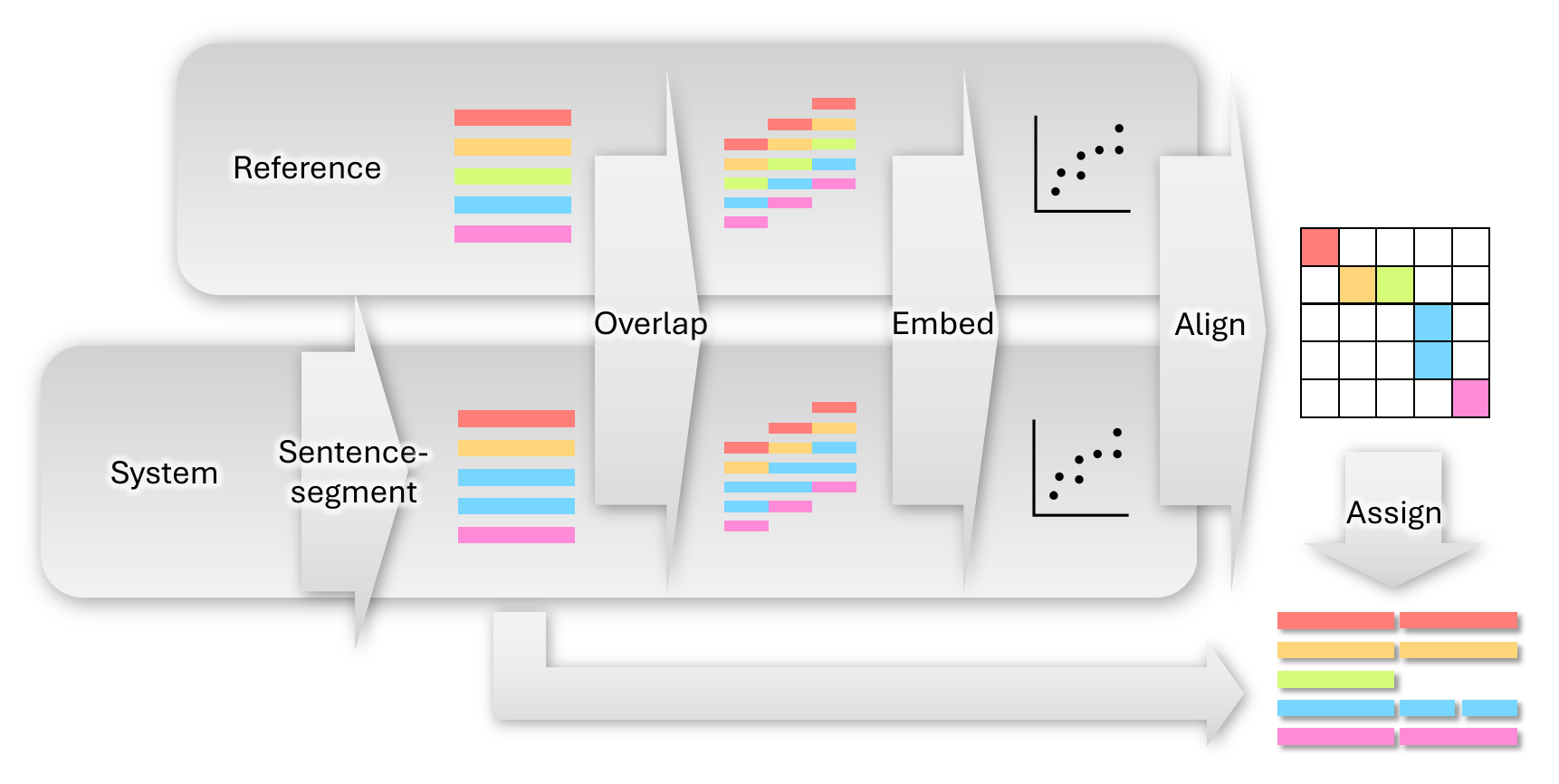}
\caption{The pipeline for automatic sentence-alignment of document-level submissions. Inputs are a PLABA reference abstract (already at sentence level) and the system output for the whole abstract. First, the system output is sentence-segmented. Then, overlapping blocks of sentences are created up to size $n$ sentences to allow 1-to-$n$ alignments. These blocks are then embedded, and the embeddings are used for scoring in the dynamic programming step. Finally, the optimal path is used to arrange system sentences (reference sentences are fixed).} \label{fig:aln}
\end{figure}

The entire pipeline for aligning sentences of document-level submissions to reference sentences is depicted in Figure~\ref{fig:aln}.
This pipeline was based on Vecalign~\citep{thompson2019vecalign}, which was developed for bilingual alignment for Machine Translation tasks but worked well in this setting. Vecalign implements efficient dynamic programming using embeddings to score similarity. Additionally, embeddings of overlapping blocks of sentences are used to allow many-to-many alignment. As in~\cite{thompson2019vecalign}, we use  LASER~\citep{artetxe2019massively} for embeddings. Initial sentence-segmentation of document-level system outputs was performed using Stanza~\citep{qi2020stanza}. Finally, since Vecalign allows insertions in one-to-many mode, but the PLABA task does not, we also developed an assignment algorithm to handle these cases.
One-to-many alignments from Vecalign allow insertions versus the source; i.e., system blocks that are marked as aligned between two source sentences, or at the beginning or end of the source. However, the PLABA task does not allow insertions, and all outputs must be assigned to a source sentence for evaluation. Any system blocks marked as insertions by the alignment process thus need to be joined to either the previous or following system block. We choose, arbitrarily, to join them to the following system block, unless the inserted block is terminal (marked as inserted after the last source sentence, with no system blocks following), in which case we join it with the previous system block. The algorithm is depicted in Algorithm~\ref{alg:asn}.

\begin{algorithm}[h!]
\caption{\centering Assignment of system sentences to reference sentences from a one-to-many alignment with possible insertions.
}
\label{alg:asn}
\begin{algorithmic}[1]
\State $\sigma$ := system sentences[\texttt{list}] \Comment{Can be modified}
\State $\alpha$ := alignment[\texttt{list} of \texttt{tuple}(src. indices[\texttt{list}], sys. indices[\texttt{list}])]
\Procedure{assign}{$\sigma,\alpha$} \Comment{Outputs string for each source sentence}
\State $\omega \gets []$ \Comment{Output list}
\For{$i$ = 1..$|\alpha|$}
\State $\iota \gets \alpha_{i,1}$ \Comment{System indices for alignment pair}
\If{$|\iota|>0$} \Comment{Pair has system indices}
\State $\zeta \gets $`' \Comment{String to join system sentences for pair}
\For{$j$ = 1..$|\iota|$}
\If{$j > 1$}
\State $\zeta \gets $concat($s, $`\ ') 
\Comment{Separate with space}
\EndIf
\State $\zeta \gets $concat($\zeta, \sigma_{\iota_j}$)
\EndFor
\If{$|\alpha_{i,0}|>0$} \Comment{Pair has source indices}
\State $\omega \gets \omega + [\zeta]$ \Comment{Append joined sentences to output list}
\Else \Comment{Insertion versus source}
\If{$i==|\alpha|$} \Comment{Terminal insertion}
\State $\omega_{|\omega|} \gets \omega_{|\omega|} + \zeta$ \Comment{Join $\zeta$ to the last output item}
\Else \Comment{Internal or leading insertion}
\State \Comment{Prepend $\zeta$ to system sentence after this aligned block}
\State $\sigma_{\iota_{|\iota|}+1} \gets \zeta + \sigma_{\iota_{|\iota|}+1}$ \EndIf
\EndIf
\Else \Comment{Deletion versus source}
\State $\omega \gets \omega + [$`'$]$  \Comment{Append blank line to output list}
\EndIf
\EndFor
\State \textbf{return} $\omega$
\EndProcedure
\end{algorithmic}
\end{algorithm}

\section{Linking Synonym Terms}\label{synonyms}
During the annotation of the Task 2 dataset, the annotators could select terms as synonyms of other terms. We developed Algorithms \ref{alg:syn_link} and \ref{alg:get_syn_set} to merge the simplifications of synonymous terms when processing the annotations for the Task 2 dataset. 

Algorithm \ref{alg:syn_link} takes two dictionaries as input: $D$, which maps terms to their simplifications, and $S$, which maps expert terms to their synonyms. The outputs of the algorithm is a new dictionary $D'$, which maps expert terms to their simplifications and the simplifications of their synonyms.

Algorithm \ref{alg:get_syn_set} is a recursive function that takes three parameters: $S$, a dictionary mapping expert terms to their synonyms, a term \textit{t} and a set of synonyms \textit{sns}. The algorithm outputs a list of all terms in \textit{S} that are synonyms with \textit{t}.

\begin{algorithm}[t]
\caption{Associate Synonyms Algorithm}\label{alg:syn_link}
\begin{algorithmic}[1]
    \Require Dictionary $D$ mapping terms to simplifications, dictionary $S$ mapping terms to synonyms
    \Ensure Dictionary $D'$ with merged synonyms
    \State $D' \gets \emptyset$
    \ForAll{$t \in S.keys()$}
        \State $sns \gets \text{get\_syn\_set($S$, $t$, $\emptyset$)}$
        \ForAll{$sn \in sns$}
            \If{$sn \neq t$}
                \ForAll{$sms \in D [t]$}
                    \State $D'[sn] \gets D[sn] \cup sms$
                \EndFor
            \EndIf
        \EndFor
    \EndFor
    \State \Return $D'$
\end{algorithmic}
\end{algorithm}

\begin{algorithm}[t]
\caption{get\_syn\_set Algorithm}\label{alg:get_syn_set}
\begin{algorithmic}[1]
    \Require Dictionary $S$ mapping terms to synonyms, term $t$, synonym set $sns$
    \Ensure Set $sns$ populated with all synonyms of $t$
    \State $sns.add(t)$
    \ForAll{$sn \in S[t]$}
        \If{$sn \neq sns$}
            \State $sns \gets \text{get\_syn\_set($S, t, sns$)}$
        \EndIf
    \EndFor
    \State \Return $sns$
\end{algorithmic}
\end{algorithm}

\section{System Descriptions for Task 1 at TREC 2023}
\label{sec:syst-task1-2023}

Team submissions to Task 1 at TREC 2023, and their descriptions, are shown in Table~\ref{tab:sys}. For the inaugural offering of PLABA, we did not know how many submissions to expect, and thus developed a handful of our own baseline systems. These explored various prompting strategies as well as both zero-shot proprietary models and fine-tuned open-source models.

\begin{figure}
    \centering
    \includegraphics[width=0.85\linewidth]{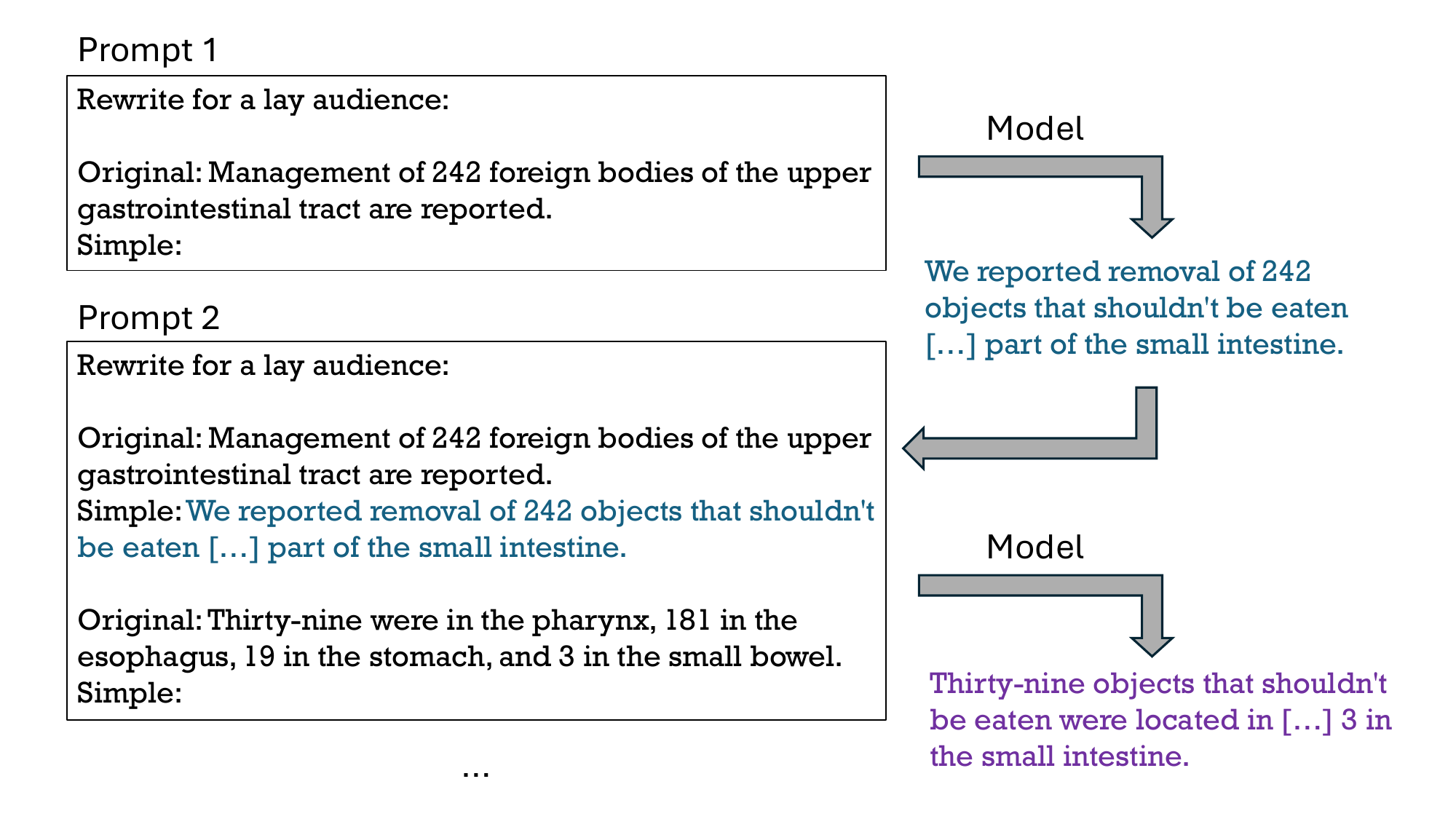}
    \caption{The sentence-wise auto-regressive zero-shot prompting strategy for the baseline system \texttt{PLABA\_1}. An initial prompt contains a general instruction (``Rewrite for a lay audience:''), the first sentence of the abstract with a label (``Original:''), and a dangling label (``Simple:''). The model's response to this prompt is then included in the next prompt, which has a dangling label for the second simple sentence. The process repeats for each sentence in the original abstract, storing model responses for each.
    }
    \label{fig:prompt}
\end{figure}

\begin{itemize}
    \item \texttt{PLABA\_1}: Sentence-level adaptation via zero-shot auto-regressive prompting of \texttt{text-davinci-003}, a variant of OpenAI's proprietary GPT 3.5 base model with instruction tuning via Reinforcement Learning from Human Feedback~\citep{ouyang2022training}. This involved progressively building prompts for each sentence of the abstract, using prior model outputs as in-context examples (Fig.~\ref{fig:prompt}). This inherently assigned outputs to individual source sentences, but also encouraged use of context when adapting each sentence. Note that, for a given source sentence, the model may return multiple sentences, essentially performing a split operation. However, these can still be attributed to a single source sentence, so alignment is not required.
    \item \texttt{PLABA\_2}: Document-level adaptation via \texttt{text-davinci-003} with instruction to rewrite each sentence (prompt: ``Rewrite each sentence for a lay audience:''). The goal of this prompting strategy was to ensure better sentence alignment, but as a document-level system.
    \item \texttt{PLABA\_3}: Document-level adaptation via zero-shot prompting of the open-source \texttt{falcon-7b-instruct} model~\citep{almazrouei2023falcon}. This baseline was included to assess how a (relatively) small, open-source language model would perform, as this type of model may be desirable for economy, privacy, and transparency. In initial tests, this model did not respond well to complex prompts such as those used for \texttt{PLABA\_1} and \texttt{PLABA\_2}; thus, the more straightforward prompt ``Simplify:'' was used, followed by the entire text of an abstract.
    \item \texttt{PLABA\_4}: \texttt{Llama-2-7B-chat} fine-tuned for the autoregressive prompting strategy of \texttt{PLABA\_1}. Training is run for 70,000 steps using a batch size of 1, Low-Rank Adaptation (LoRA)~\citep{hu2022lora} with r = 16 and $\alpha=32$, and a maximum sequence length of 4,096. Source-Copying Exposure Regularization (SCER)~\citep{ondov2024sentence} is employed to encourage simplification over source-copying.
    \item \texttt{PLABA\_5}: The same model as \texttt{PLABA\_4}, but fine-tuned without SCER.
\end{itemize}

\begin{table*}[h!]
\caption{
    Submissions to the first offering of Task 1 at TREC 2023. Each submission is numbered by team-specified priority for evaluation (lower number is higher priority). System descriptions are as provided by teams.}
    \label{tab:sys}
    \centering
    \small
    \begin{tabularx}{\textwidth}{p{0.15\linewidth}p{0.75\linewidth}}
    \textbf{Submission} & \textbf{Description} \\
    \toprule
    \toprule
\multicolumn{2}{c}{\makecell{\textbf{BoschAI (Bosch Center for Artificial Intelligence)} \\ \citep{knappich2023boschai}}} \\
\hline
BoschAI\_1 & ``LLama 13B trained with token-based weight losses derived from diffs between source and target. For every sentence, a simplification is generated with varying temperatures. The best simplification is then selected using ChatGPT.'' \\
BoschAI\_2 & ``LLama 13B trained with token-based weight losses derived from diffs between source and target.'' \\
BoschAI\_3 & ``Zero-shot prompting of ChatGPT in a sentence-by-sentence fashion.'' \\
\midrule
\multicolumn{2}{c}{\textbf{MasonNLP (George Mason University)} \citep{gangavarapu2025adapting}} \\
\hline
MasonNLP\_1 & ``One shot, instruction tuning on GPT-4. Instructions were custom-designed for this task. This had a top preference in our human evaluation.'' \\
MasonNLP\_2 & ``Predictions from a GPT3.5 instruction-tuned model on the training set and had a second preference in our human evaluation.'' \\
MasonNLP\_3 & ``Predictions from the model LLaMa13B-chat fine-tuned on the PLABA training set. This model had 3rd preference in our human evaluation.'' \\
\midrule
\multicolumn{2}{c}{\textbf{BeeManc (University of Manchester)} \citep{li2024beemanc,li2024large}} \\
\hline
BeeManc\_1 & ``Bart-large with control tokens'' \\
BeeManc\_2 & ``T5'' \\
BeeManc\_3 & ``sci-T5'' \\
\midrule
\multicolumn{2}{c}{\textbf{\makecell{PT3M (Palestine Technical University – Kadoorie,\\ Universidad Carlos III de Madrid)}}} \\
\hline
PT3M\_1* & ``Zero-shot adaptation for biomedical abstracts with biobart model version 2'' \\
\bottomrule
\bottomrule
\multicolumn{2}{l}{\footnotesize{\makecell[l]{*Automatically aligned document-level submission (see \S\ref{sec:aln}).}}}
\end{tabularx}
\end{table*}

\section{System Descriptions for Task 1 at TREC 2024}
\label{sec:syst-task1-2024}

Team submissions to Task 1 at TREC 2024, and their metadata, are shown in Figures \ref{syst-task1-BU}-\ref{syst-task1-Yseop}.

\begin{figure*}
\small
	   
\begin{tcolorbox}[colback=gray!5, colframe=black, title=BU\_1 (Task 1 at TREC 2024)]
\textbf{Full name}: LLaMA-8B-4bit-MedicalAbstract-seq-to-seq-v1\\
\textbf{Base model(s)}: Meta-Llama-3.1-8B-instruct\\
\textbf{Manual intervention}: -\\
\textbf{Training data}: Task 2 data\\
\textbf{Description}: ``What sets this project apart is its use of 4-bit quantization, which makes the model more efficient in terms of memory usage. The model was also trained using a custom dictionary to help explain complex medical terms, which makes the output much clearer and more accessible to non-medical professionals. Also, introduced while configuring the tokenizer - padding side - to help sentence length match up i.e. input and output sentence structure similarity to optimize the output.''
\end{tcolorbox}

\caption{Metadata for the \texttt{BU} submissions to Task 1 at TREC 2024, by team-selected priority. Full names and quoted descriptions are verbatim; other responses are edited for consistency of formatting and model naming.}
\label{syst-task1-BU}

\end{figure*}

\begin{figure*}
\small
	   
\begin{tcolorbox}[colback=gray!5, colframe=black, title=CLAC\_1 (Task 1 at TREC 2024)]
\textbf{Full name}: mistral-FINAL\\
\textbf{Base model(s)}: mistral-large-latest\\
\textbf{Manual intervention}: Deletion of quotes\\
\textbf{Training data}: Task 2 data\\
\textbf{Description}: ``mistral-large-latest 7-shot with temperature 0.4''
\end{tcolorbox}
\begin{tcolorbox}[colback=gray!5, colframe=black, title=CLAC\_2 (Task 1 at TREC 2024)]
\textbf{Full name}: mistral-fix\\
\textbf{Base model(s)}: mistral-large-latest\\
\textbf{Manual intervention}: Deletion of quotes; fixing q34\\
\textbf{Training data}: Task 2 data\\
\textbf{Description}: ``mistral-large-latest 7-shot with temp 0.4''
\end{tcolorbox}
\begin{tcolorbox}[colback=gray!5, colframe=black, title=CLAC\_3 (Task 1 at TREC 2024)]
\textbf{Full name}: gpt-final\\
\textbf{Base model(s)}: gpt-3.5-turbo-0125\\
\textbf{Manual intervention}: Deletion of quotes\\
\textbf{Training data}: Task 2 data\\
\textbf{Description}: ``gpt-3.5-turbo-0125 7-shot run''
\end{tcolorbox}

\caption{Metadata for the \texttt{CLAC} submissions to Task 1 at TREC 2024, by team-selected priority. Full names and quoted descriptions are verbatim; other responses are edited for consistency of formatting and model naming.}
\label{syst-task1-CLAC}

\end{figure*}

\begin{figure*}
\small
	   
\begin{tcolorbox}[colback=gray!5, colframe=black, title=ntu\_nlp\_1 (Task 1 at TREC 2024)]
\textbf{Full name}: task2\_moa\_tier1\_post\\
\textbf{Base model(s)}: gemini-1.0-pro, gemini-1.5-flash, gemini-1.5-pro, gemma-2-27b, gpt-4o-mini, Meta-Llama-3.1-8B, Mistral-Nemo-Instruct-2407\\
\textbf{Manual intervention}: -\\
\textbf{Training data}: PLABA dataset (Attal 2023)\\
\textbf{Description}: ``Method: demonstration relevant 5 shot + NER None \& Finetune 1epoch NER None''
\end{tcolorbox}
\begin{tcolorbox}[colback=gray!5, colframe=black, title=ntu\_nlp\_2 (Task 1 at TREC 2024)]
\textbf{Full name}: task2\_moa\_tier2\_post\\
\textbf{Base model(s)}: gemini-1.0-pro, gemini-1.5-flash, gemini-1.5-pro, gemma-2-27b, gpt-4o-mini, Meta-Llama-3.1-8B, Mistral-Nemo-Instruct-2407\\
\textbf{Manual intervention}: -\\
\textbf{Training data}: PLABA dataset (Attal 2023)\\
\textbf{Description}: ``Method: demonstration relevant 5 shot + NER None''
\end{tcolorbox}
\begin{tcolorbox}[colback=gray!5, colframe=black, title=ntu\_nlp\_3 (Task 1 at TREC 2024)]
\textbf{Full name}: task2\_moa\_tier3\_post\\
\textbf{Base model(s)}: gemini-1.0-pro, gemini-1.5-flash, gemini-1.5-pro, gemma-2-27b, gpt-4o-mini, Meta-Llama-3.1-8B, Mistral-Nemo-Instruct-2407\\
\textbf{Manual intervention}: -\\
\textbf{Training data}: PLABA dataset (Attal 2023)\\
\textbf{Description}: ``Method: demonstration relevant 5 shot + NER 5''
\end{tcolorbox}

\caption{Metadata for the \texttt{ntu\_nlp} submissions to Task 1 at TREC 2024, by team-selected priority. Full names and quoted descriptions are verbatim; other responses are edited for consistency of formatting and model naming.}
\label{syst-task1-ntunlp}

\end{figure*}

\begin{figure*}
\small
	   
\begin{tcolorbox}[colback=gray!5, colframe=black, title=SIB\_1 (Task 1 at TREC 2024)]
\textbf{Full name}: TREC2024\_SIB\_run3\\
\textbf{Base model(s)}: Meta-Llama-3.1\\
\textbf{Manual intervention}: -\\
\textbf{Training data}: -\\
\textbf{Description}: ``Basic prompting with Llama3.1 instead of Llama3''
\end{tcolorbox}
\begin{tcolorbox}[colback=gray!5, colframe=black, title=SIB\_2 (Task 1 at TREC 2024)]
\textbf{Full name}: TREC2024\_SIB\_run1\\
\textbf{Base model(s)}: Meta-Llama-3-8B\\
\textbf{Manual intervention}: -\\
\textbf{Training data}: PLABA dataset (Attal 2023)\\
\textbf{Description}: ``This run is a baseline, with the original model (no special feature), to compare the performance of our other runs''
\end{tcolorbox}
\begin{tcolorbox}[colback=gray!5, colframe=black, title=SIB\_3 (Task 1 at TREC 2024)]
\textbf{Full name}: TREC2024\_SIB\_run4\\
\textbf{Base model(s)}: Meta-Llama-3-8B\\
\textbf{Manual intervention}: -\\
\textbf{Training data}: -\\
\textbf{Description}: ``Retrieval Augmented Generation (RAG) based on documents from Wikipedia and Monash''
\end{tcolorbox}

\caption{Metadata for the \texttt{SIB} submissions to Task 1 at TREC 2024, by team-selected priority. Full names and quoted descriptions are verbatim; other responses are edited for consistency of formatting and model naming.}
\label{syst-task1-SIB}

\end{figure*}

\begin{figure*}
\small
	   
\begin{tcolorbox}[colback=gray!5, colframe=black, title=UAmsterdam\_1 (Task 1 at TREC 2024)]
\textbf{Full name}: UAms-ConBART-Cochrane\\
\textbf{Base model(s)}: BART\\
\textbf{Manual intervention}: -\\
\textbf{Training data}: Cochrane plain English summaries\\
\textbf{Description}: ``Contextual BART model trained as planned guided simplification model, using with a ``rephrase" instruction for each sentence.''
\end{tcolorbox}
\begin{tcolorbox}[colback=gray!5, colframe=black, title=UAmsterdam\_2 (Task 1 at TREC 2024)]
\textbf{Full name}: UAms-BART-Cochrane\\
\textbf{Base model(s)}: BART\\
\textbf{Manual intervention}: -\\
\textbf{Training data}: Cochrane plain English summaries\\
\textbf{Description}: ``Sentence BART model trained as planned guided simplification model, using with a ``rephrase" instruction for each sentence.''
\end{tcolorbox}

\caption{Metadata for the \texttt{UAmsterdam} submissions to Task 1 at TREC 2024, by team-selected priority. Full names and quoted descriptions are verbatim; other responses are edited for consistency of formatting and model naming.}
\label{syst-task1-UAmsterdam}

\end{figure*}

\begin{figure*}
\small
	   
\begin{tcolorbox}[colback=gray!5, colframe=black, title=UM\_1 (Task 1 at TREC 2024)]
\textbf{Full name}: LLaMa 3.1 70B instruction (2nd run)\\
\textbf{Base model(s)}: Meta-Llama-3.1-70B-instruct\\
\textbf{Manual intervention}: -\\
\textbf{Training data}: -\\
\textbf{Description}: ``We used transformers.pipeline, with all default args''
\end{tcolorbox}
\begin{tcolorbox}[colback=gray!5, colframe=black, title=UM\_2 (Task 1 at TREC 2024)]
\textbf{Full name}: GPT\\
\textbf{Base model(s)}: gpt-4o\\
\textbf{Manual intervention}: Alignment\\
\textbf{Training data}: PLABA dataset (Attal 2023)\\
\textbf{Description}: ``Self-prompt until the output is satisfied.''
\end{tcolorbox}

\caption{Metadata for the \texttt{UM} submissions to Task 1 at TREC 2024, by team-selected priority \citep{ling2025maleiplabatracktrec}. Full names and quoted descriptions are verbatim; other responses are edited for consistency of formatting and model naming.}
\label{syst-task1-UM}

\end{figure*}

\begin{figure*}
\small
	   
\begin{tcolorbox}[colback=gray!5, colframe=black, title=um\_fhs\_1 (Task 1 at TREC 2024)]
\textbf{Full name}: plaba\_um\_fhs\_sub1\\
\textbf{Base model(s)}: gpt-4o-mini-2024-07-18\\
\textbf{Manual intervention}: -\\
\textbf{Training data}: -\\
\textbf{Description}: ``Two GPT-4o mini, model version gpt-4o-mini-2024-07-18, AI agent were created, where after creating the adaptation with the AI agent 1, AI agents 2 in the persona of a ``smart 13-14 year old student" who asked clarification questions and then the adaptation was modified using relevant answers to the questions.''
\end{tcolorbox}
\begin{tcolorbox}[colback=gray!5, colframe=black, title=um\_fhs\_2 (Task 1 at TREC 2024)]
\textbf{Full name}: plaba\_um\_fhs\_sub2\\
\textbf{Base model(s)}: gpt-4o-mini-2024-07-18\\
\textbf{Manual intervention}: -\\
\textbf{Training data}: -\\
\textbf{Description}: ``Two GPT-4o mini, model version gpt-4o-mini-2024-07-18, with a user prompt including the manually adapted guidelines was used.''
\end{tcolorbox}
\begin{tcolorbox}[colback=gray!5, colframe=black, title=um\_fhs\_3 (Task 1 at TREC 2024)]
\textbf{Full name}: plaba\_um\_fhs\_sub3\\
\textbf{Base model(s)}: gpt-4o-2024-08-06\\
\textbf{Manual intervention}: -\\
\textbf{Training data}: PLABA dataset (Attal 2023)\\
\textbf{Description}: ``GPT-4o, model version gpt-4o-2024-08-06, was  fine-tuned on the PLABA dataset (80\% for training and 20\% for validation) was used for this run.''
\end{tcolorbox}

\caption{Metadata for the \texttt{um\_fhs} submissions to Task 1 at TREC 2024, by team-selected priority \citep{kocbek2025um_fhs}. Full names and quoted descriptions are verbatim; other responses are edited for consistency of formatting and model naming.}
\label{syst-task1-umfhs}

\end{figure*}

\begin{figure*}
\small
	   
\begin{tcolorbox}[colback=gray!5, colframe=black, title=Yseop\_1 (Task 1 at TREC 2024)]
\textbf{Full name}: gpt35\_dspy\\
\textbf{Base model(s)}: gpt-35-turbo-16k\\
\textbf{Manual intervention}: -\\
\textbf{Training data}: PLABA dataset (Attal 2023)\\
\textbf{Description}: ``prompt engg using dspy''
\end{tcolorbox}
\begin{tcolorbox}[colback=gray!5, colframe=black, title=Yseop\_2 (Task 1 at TREC 2024)]
\textbf{Full name}: bart\_base\_ft\\
\textbf{Base model(s)}: facebook/bart-base\\
\textbf{Manual intervention}: -\\
\textbf{Training data}: PLABA dataset (Attal 2023)\\
\textbf{Description}: ``high sari score but mediocre quality sentences''
\end{tcolorbox}

\caption{Metadata for the \texttt{Yseop} submissions to Task 1 at TREC 2024, by team-selected priority. Full names and quoted descriptions are verbatim; other responses are edited for consistency of formatting and model naming.}
\label{syst-task1-Yseop}

\end{figure*}

\section{System Descriptions for Task 2 at TREC 2024}
\label{sec:syst-task2-2024}

Team submissions to Task 1 at TREC 2024, and their metadata, are shown in Figures \ref{syst-task2-BU}-\ref{syst-task2-Yseop}.

We also developed baseline systems for the identification (Task 2A) and classification (Task 2B) sub-tasks within Task 2.
This included three baselines for Task 2A:
\begin{itemize}
    \item \texttt{PLABA\_2A\_1}: Uses MetaMapLite \citep{demner2017metamap} and the Unified Medical Language System (UMLS) \citep{lindberg1993unified} to identify expert terms. Two term frequency datasets from Kaggle---one derived from the Google Web Trillion Word Corpus \cite{unigrams} and the other derived from BookCorpus and a 2019 dump of Wikipedia \cite{bigrams}---were used to filter out false positives.
    \item \texttt{PLABA\_2A\_2}: DeBERTa Large \citep{deberta} (435M parameters) was fine-tuned for the identification task by treating it as a named entity recognition problem.
    \item \texttt{PLABA\_2A\_3}: Llama3 Instruct (8B parameters) \citep{dubey2024llama} was prompted for this task by providing the following instruction prompt immediately followed by the sentence to operate on: ``Identify all non-consumer biomedical terms in the user's sentence using a comma-separated list. Generate no other text besides the list.''
\end{itemize}

We developed two baselines for Task 2B, which also required performing Task 2A as an intermediate step:
\begin{itemize}
    \item \texttt{PLABA\_2B\_1}: BERT Large (340M parameters) \citep{bert} was fine-tuned for a named entity recognition problem that combined the identification and classification sub-tasks into a single problem.
    \item \texttt{PLABA\_2B\_2}: DeBERTa Large was fine-tuned for the same named entity recognition problem as \texttt{CLS-BERT}.
\end{itemize}

\begin{figure*}
\small
	   
\begin{tcolorbox}[colback=gray!5, colframe=black, title=BU\_1 (Task 2 at TREC 2024)]
\textbf{Full name}: MLPClassifier-identify-classify-replace-v1\\
\textbf{Base-LLM(s)}: -\\
\textbf{Manual intervention}: -\\
\textbf{Training data}: -\\
\textbf{Description}: ``This run explores different classifiers (XGBoost, LightGBM)and uses an MLP, which can capture non-linear patterns. The model accuracy is about 65\%. However, a deep dive into the performance of individual action shows the F1 for each class''
\end{tcolorbox}

\caption{Metadata for the \texttt{BU} submissions to Task 2 at TREC 2024, by team-selected priority. Full names and quoted descriptions are verbatim; other responses are edited for consistency of formatting and model naming.}
\label{syst-task2-BU}

\end{figure*}

\begin{figure*}
\small
	   
\begin{tcolorbox}[colback=gray!5, colframe=black, title=CLAC\_1 (Task 2 at TREC 2024)]
\textbf{Full name}: mistral\\
\textbf{Base-LLM(s)}: mistral-large-latest\\
\textbf{Manual intervention}: -\\
\textbf{Training data}: -\\
\textbf{Description}: ``mistral-large-latest 7-shot with temperature of 0.4''
\end{tcolorbox}
\begin{tcolorbox}[colback=gray!5, colframe=black, title=CLAC\_2 (Task 2 at TREC 2024)]
\textbf{Full name}: gpt\\
\textbf{Base-LLM(s)}: gpt-3.5-turbo-0125\\
\textbf{Manual intervention}: -\\
\textbf{Training data}: Task 2 training set\\
\textbf{Description}: ``gpt-3.5-turbo-0125 7-shot run''
\end{tcolorbox}

\caption{Metadata for the \texttt{CLAC} submissions to Task 2 at TREC 2024, by team-selected priority. Full names and quoted descriptions are verbatim; other responses are edited for consistency of formatting and model naming.}
\label{syst-task2-CLAC}

\end{figure*}

\begin{figure*}
\small
	   
\begin{tcolorbox}[colback=gray!5, colframe=black, title=IIITH\_1 (Task 2 at TREC 2024)]
\textbf{Full name}: First\\
\textbf{Base-LLM(s)}: BioBERT\\
\textbf{Manual intervention}: Formatting\\
\textbf{Training data}: Task 2 training set\\
\textbf{Description}: ``The following run is computationally cheap as it does not require the use of any LLMs. The given data was also complimented further using other publicly available datasets''
\end{tcolorbox}

\caption{Metadata for the \texttt{IIITH} submissions to Task 2 at TREC 2024, by team-selected priority. Full names and quoted descriptions are verbatim; other responses are edited for consistency of formatting and model naming.}
\label{syst-task2-IIITH}

\end{figure*}

\begin{figure*}
\small
	   
\begin{tcolorbox}[colback=gray!5, colframe=black, title=ntu\_nlp\_1 (Task 2 at TREC 2024)]
\textbf{Full name}: gemini-1.5-pro\_demon5\_replace-demon5\\
\textbf{Base-LLM(s)}: gemini-pro-1.5\\
\textbf{Manual intervention}: Formatting\\
\textbf{Training data}: -\\
\textbf{Description}: ``Use gemini-pro-1.5 as base model.''
\end{tcolorbox}
\begin{tcolorbox}[colback=gray!5, colframe=black, title=ntu\_nlp\_2 (Task 2 at TREC 2024)]
\textbf{Full name}: gemini-1.5-flash\_demon5\_replace-demon5\\
\textbf{Base-LLM(s)}: gemini-1.5-flash\\
\textbf{Manual intervention}: Formatting\\
\textbf{Training data}: -\\
\textbf{Description}: ``Use gemini-flash-1.5 as base model.''
\end{tcolorbox}
\begin{tcolorbox}[colback=gray!5, colframe=black, title=ntu\_nlp\_3 (Task 2 at TREC 2024)]
\textbf{Full name}: gpt-4o-mini \_demon5\_replace-demon5\\
\textbf{Base-LLM(s)}: gpt-4o-mini\\
\textbf{Manual intervention}: Formatting\\
\textbf{Training data}: -\\
\textbf{Description}: ``Use gpt-4o-mini as base model.''
\end{tcolorbox}

\caption{Metadata for the \texttt{ntu\_nlp} submissions to Task 2 at TREC 2024, by team-selected priority. Full names and quoted descriptions are verbatim; other responses are edited for consistency of formatting and model naming.}
\label{syst-task2-ntunlp}

\end{figure*}

\begin{figure*}
\small
	   
\begin{tcolorbox}[colback=gray!5, colframe=black, title=UM\_1 (Task 2 at TREC 2024)]
\textbf{Full name}: Roberta-base\\
\textbf{Base-LLM(s)}: RoBERTa-base\\
\textbf{Manual intervention}: -\\
\textbf{Training data}: -\\
\textbf{Description}: ``Multi label token classification with roberta-base''
\end{tcolorbox}

\caption{Metadata for the \texttt{UM} submissions to Task 2 at TREC 2024, by team-selected priority. Full names and quoted descriptions are verbatim; other responses are edited for consistency of formatting and model naming.}
\label{syst-task2-UM}

\end{figure*}

\begin{figure*}
\small
	   
\begin{tcolorbox}[colback=gray!5, colframe=black, title=Yseop\_1 (Task 2 at TREC 2024)]
\textbf{Full name}: roberta-gbc\\
\textbf{Base-LLM(s)}: pabRomero/BioMedRoBERTa-full-finetuned-ner-pablo\\
\textbf{Manual intervention}: -\\
\textbf{Training data}: -\\
\textbf{Description}: ``cleaned the training corpus and hyper param tuning of the two models''
\end{tcolorbox}

\caption{Metadata for the \texttt{Yseop} submissions to Task 2 at TREC 2024, by team-selected priority. Full names and quoted descriptions are verbatim; other responses are edited for consistency of formatting and model naming.}
\label{syst-task2-Yseop}

\end{figure*}

\section{Inter-annotator Agreement for Manual Evaluation of Task 1}
\label{apx:iaa}

For the second round of manual annotation, resources also allowed for giving annotators overlapping assignments for the three abstracts in order to estimate inter-annotator agreement. Results for each axis are shown in Table~\ref{tab:agr}. 
Agreement is generally high, with two notable exceptions being sentence simplicity (SEN) and fluency (FLU). Low agreement for fluency is due to over 98\% of judgments being 1 (perfect), which breaks distributive assumptions of $\alpha$. On inspection of differences in sentence simplicity, annotators disagree on how to rate outputs that are too long but with each sentence being relatively short. We address this in the subsequent offering by providing a ``brevity'' axis and clarifying guidelines. Future tasks should continue to make such distinctions explicit in annotation guidelines.

\begin{table}
\centering
\small
\caption{Inter-annotator agreement for three abstracts, via Krippendorf's $\alpha$, and at different levels of aggregation (sentence level being the original annotations).}\label{tab:agr}
\begin{tabular}{lcccccc}
\hline
\textbf{Aggregation} & \textbf{SEN} & \textbf{TRM} & \textbf{TAC} & \textbf{FLU} & \textbf{COM} & \textbf{FTH}\\
\hline
\textbf{Sentence} & 0.1748 & 0.5279 & 0.5845 & -0.0255 & 0.8833 & 0.8148\\
\textbf{Abstract} & -0.1311 & 0.854 & 0.6757 & 0.3241 & 0.9684 & 0.8823\\
\textbf{System} & 0.0769 & 0.8077 & 0.6026 & 0.0225 & 0.9853 & 0.7543\\
\end{tabular}
\end{table}

\section{Scatterplots of Automatic Metrics Versus Manual Judgments of Task 1 at TREC 2023}
\label{apx:met}

\begin{figure}
     \centering
     \begin{subfigure}[b]{0.48\textwidth}
         \centering
         \includegraphics[width=\textwidth]{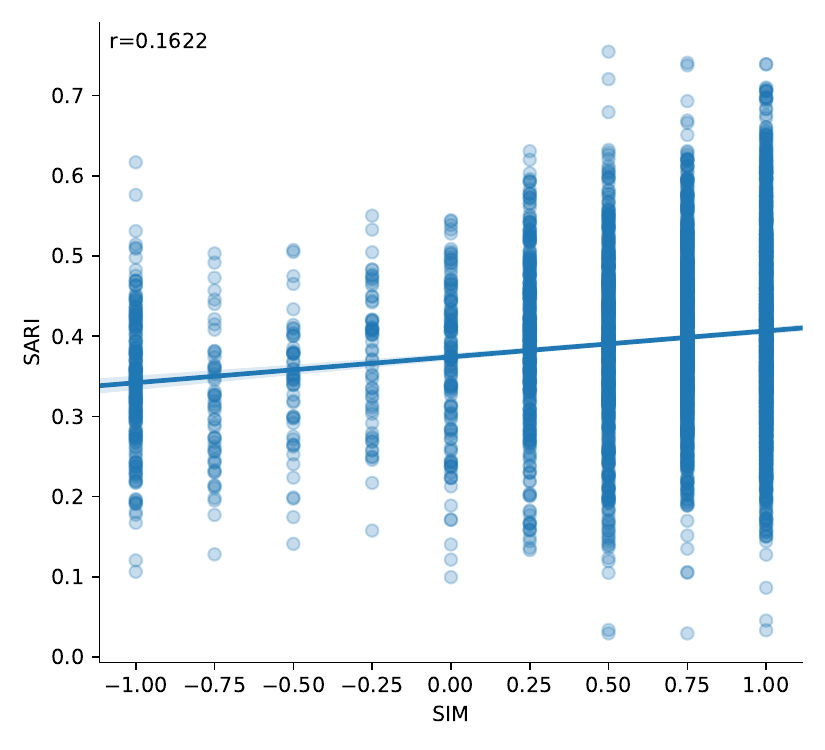}
     \end{subfigure}
     \hfill
     \begin{subfigure}[b]{0.48\textwidth}
         \centering
         \includegraphics[width=\textwidth]{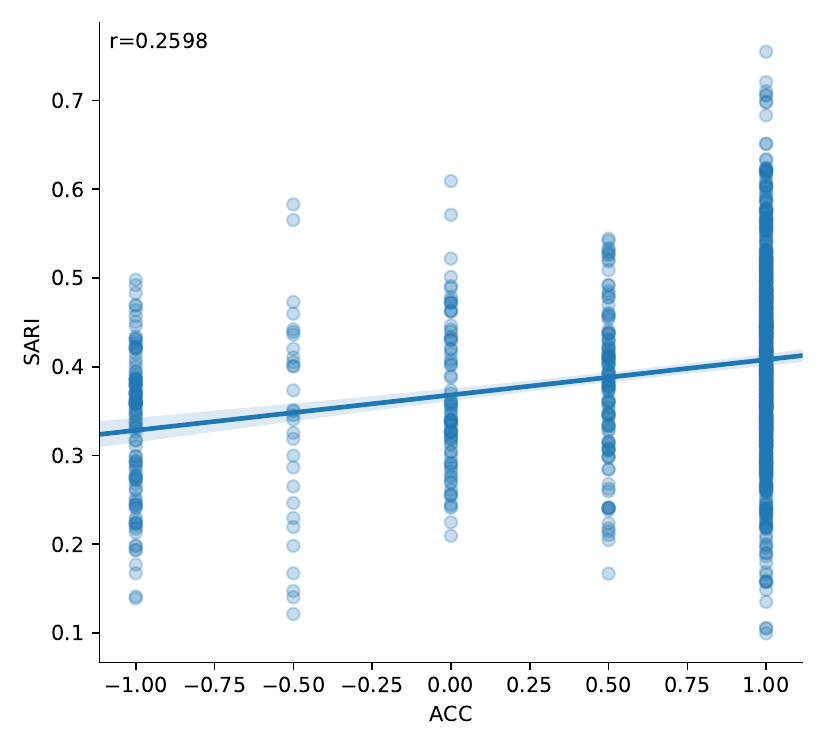}
     \end{subfigure}
     \begin{subfigure}[b]{0.48\textwidth}
         \centering
         \includegraphics[width=\textwidth]{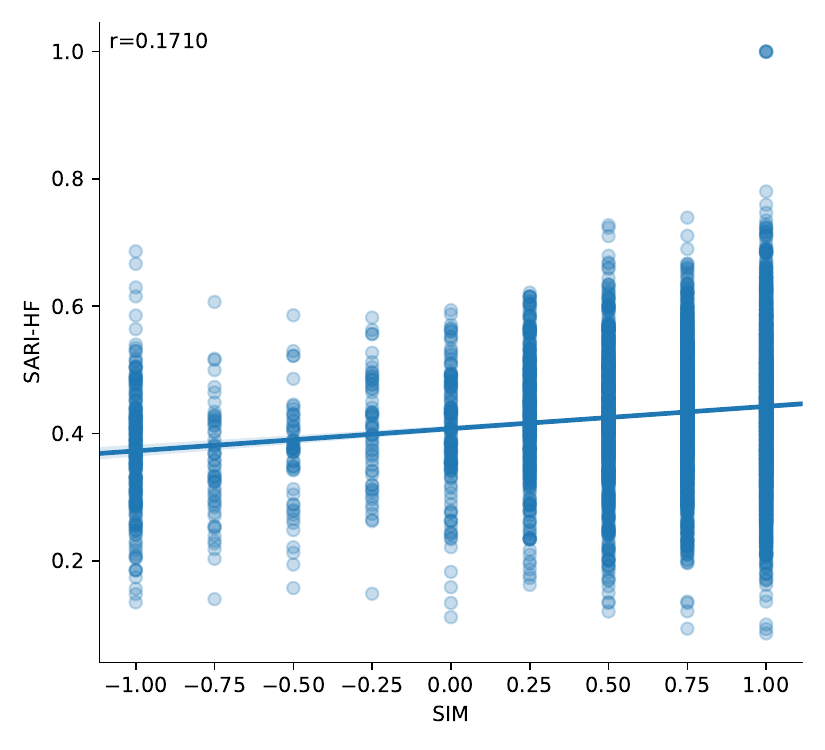}
     \end{subfigure}
     \hfill
     \begin{subfigure}[b]{0.48\textwidth}
         \centering
         \includegraphics[width=\textwidth]{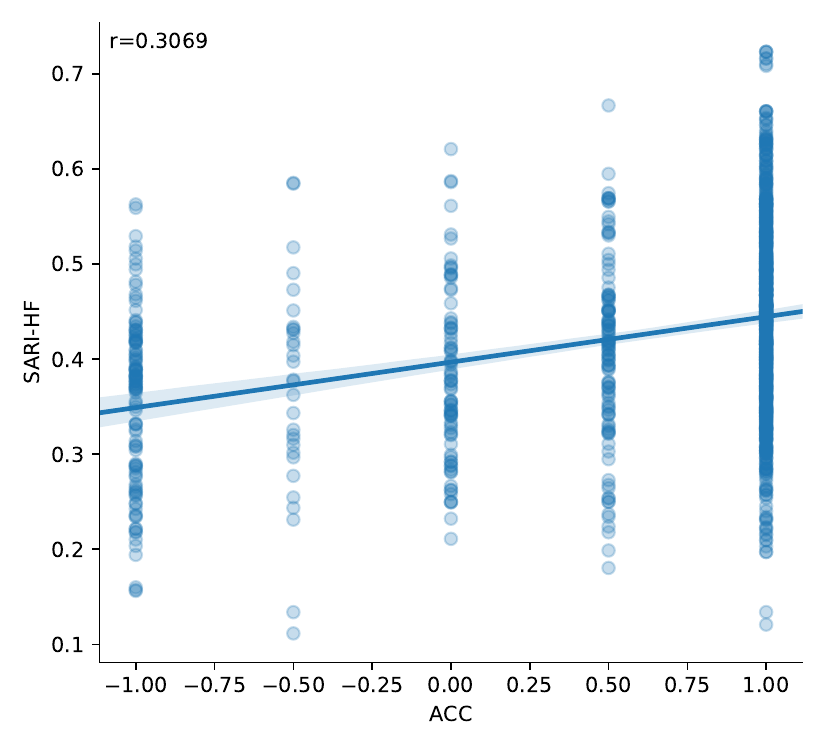}
     \end{subfigure}
     \caption{Scatterplots and regression lines of \texttt{SARI} (top, original implementation; bottom, HuggingFace implemenation) versus manual judgments of simplicity (SIM, left) and accuracy (ACC, right).}
    \label{fig:hug}
\end{figure}

\begin{figure}
     \begin{subfigure}[b]{0.48\textwidth}
         \centering
         \includegraphics[width=\textwidth]{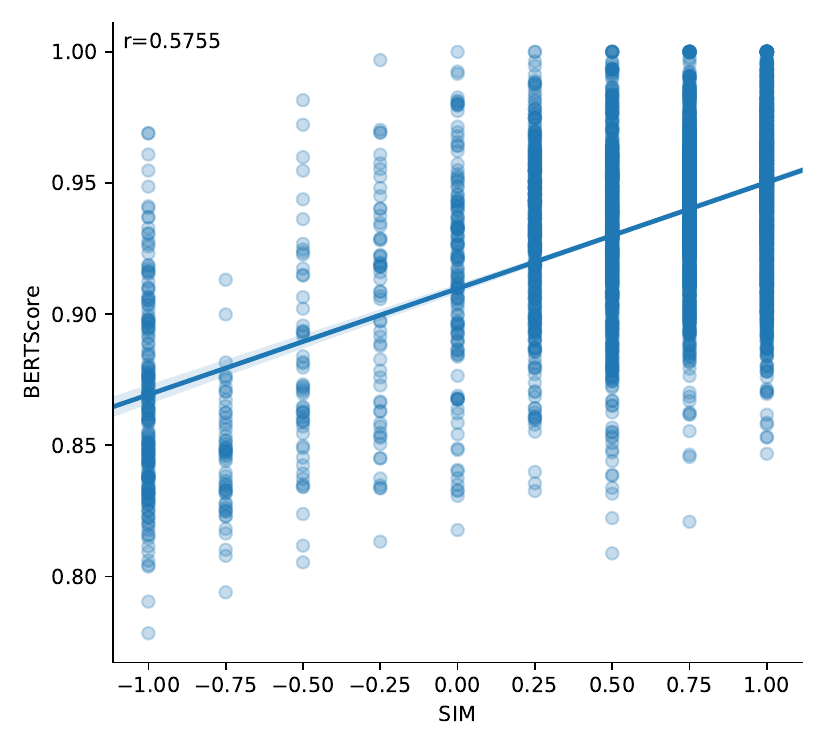}
     \end{subfigure}
     \hfill
     \begin{subfigure}[b]{0.48\textwidth}
         \centering
         \includegraphics[width=\textwidth]{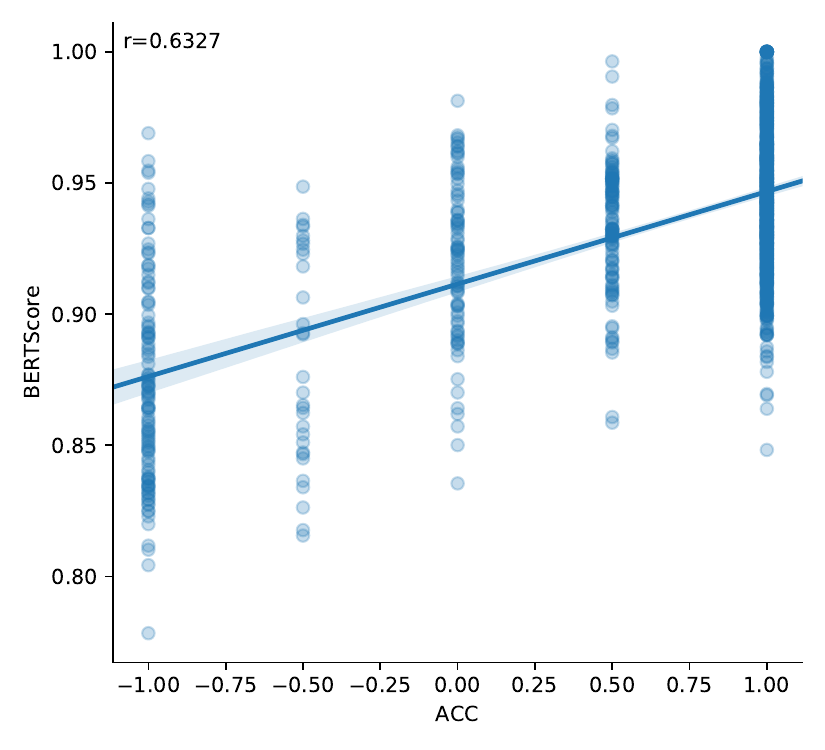}
     \end{subfigure}
     \caption{Scatterplots and regression lines of \texttt{BERTScore} versus manual judgments of simplicity (SIM, left) and accuracy (ACC, right).}
    \label{fig:bert}
\end{figure}

\begin{figure}
     \begin{subfigure}[b]{0.48\textwidth}
         \centering
         \includegraphics[width=\textwidth]{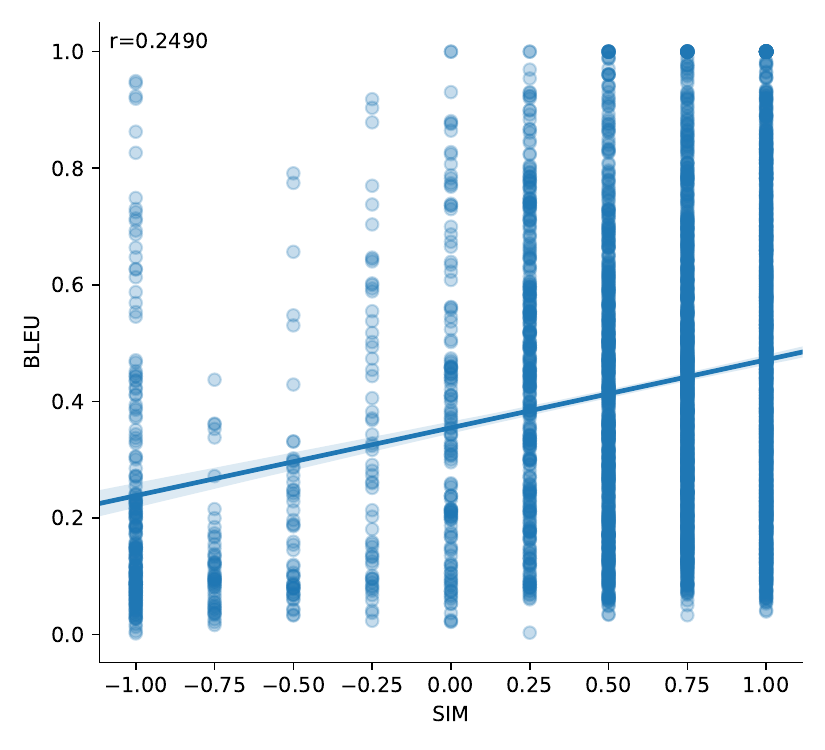}
     \end{subfigure}
     \hfill
     \begin{subfigure}[b]{0.48\textwidth}
         \centering
         \includegraphics[width=\textwidth]{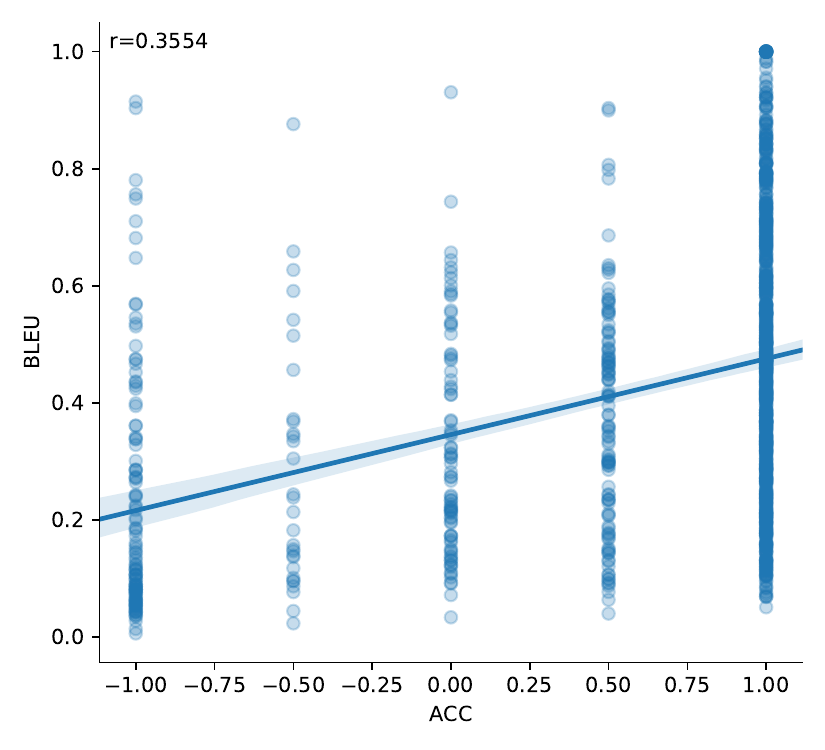}
     \end{subfigure}
        \caption{Scatterplots and regression lines of \texttt{BLEU} versus manual judgments of simplicity (SIM, left) and accuracy (ACC, right).}
    \label{fig:blu}
\end{figure}

\begin{figure}
     \begin{subfigure}[b]{0.48\textwidth}
         \centering
    \includegraphics[width=\textwidth]{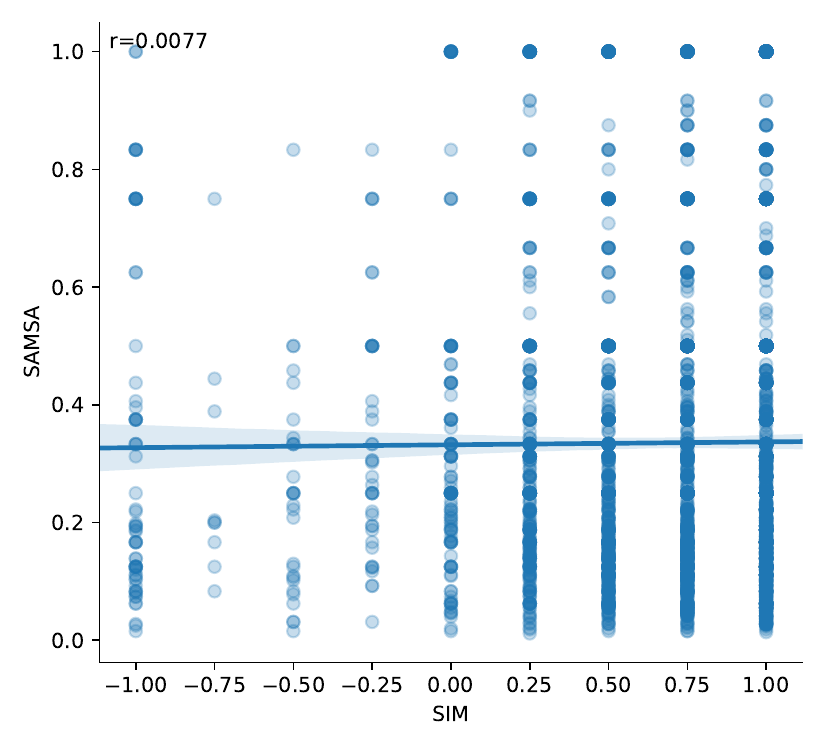}
     \end{subfigure}
     \hfill
     \begin{subfigure}[b]{0.48\textwidth}
         \centering
         \includegraphics[width=\textwidth]{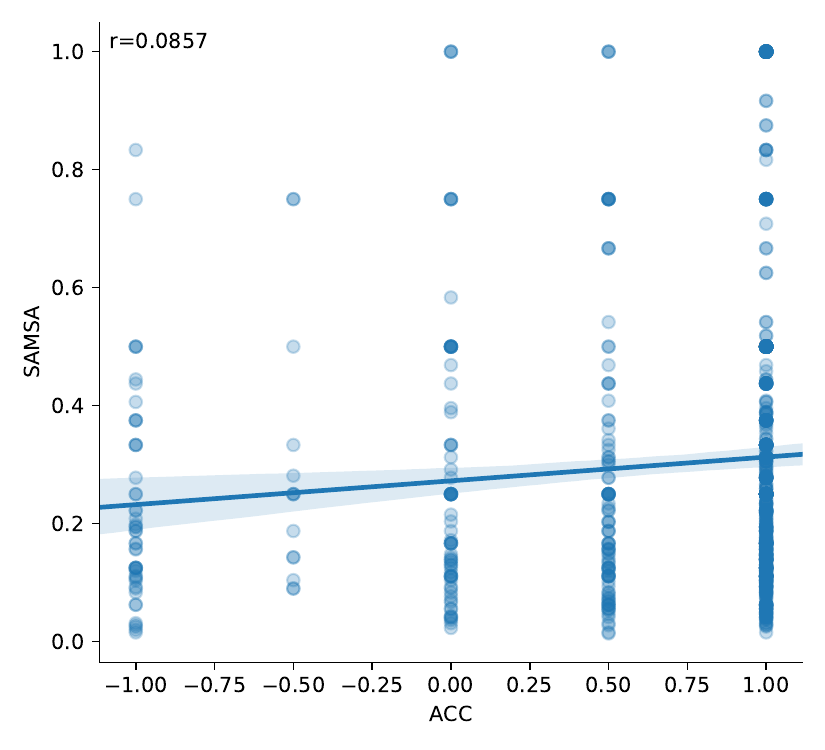}
     \end{subfigure}
        \caption{Scatterplots and regression lines of \texttt{SAMSA} versus manual judgments of simplicity (SIM, left) and accuracy (ACC, right).}
    \label{fig:sam}
\end{figure}

\begin{figure}
     \centering
     \begin{subfigure}[b]{0.48\textwidth}
         \centering
         \includegraphics[width=\textwidth]{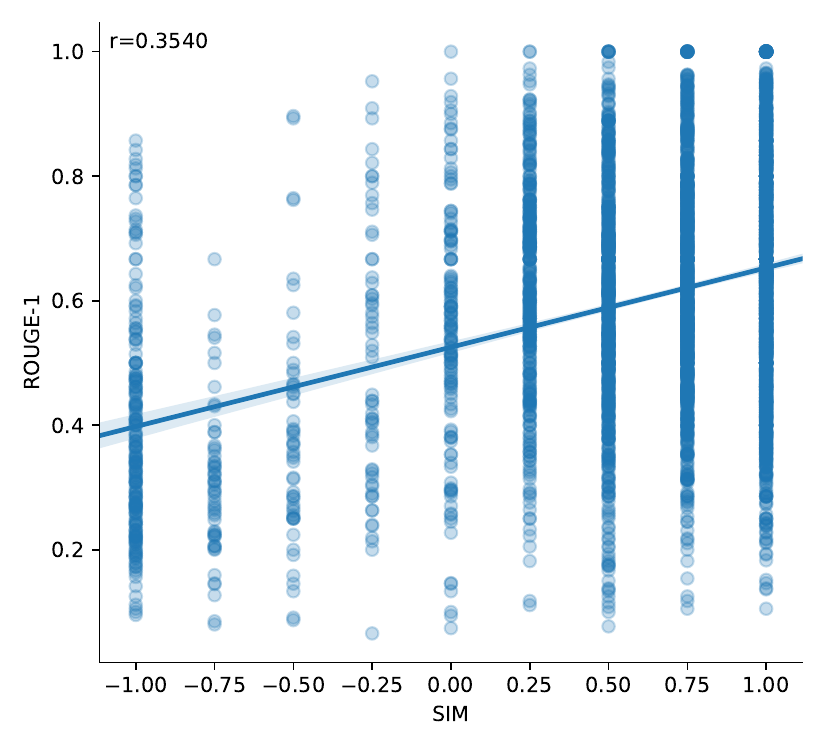}
     \end{subfigure}
     \hfill
     \begin{subfigure}[b]{0.48\textwidth}
         \centering
         \includegraphics[width=\textwidth]{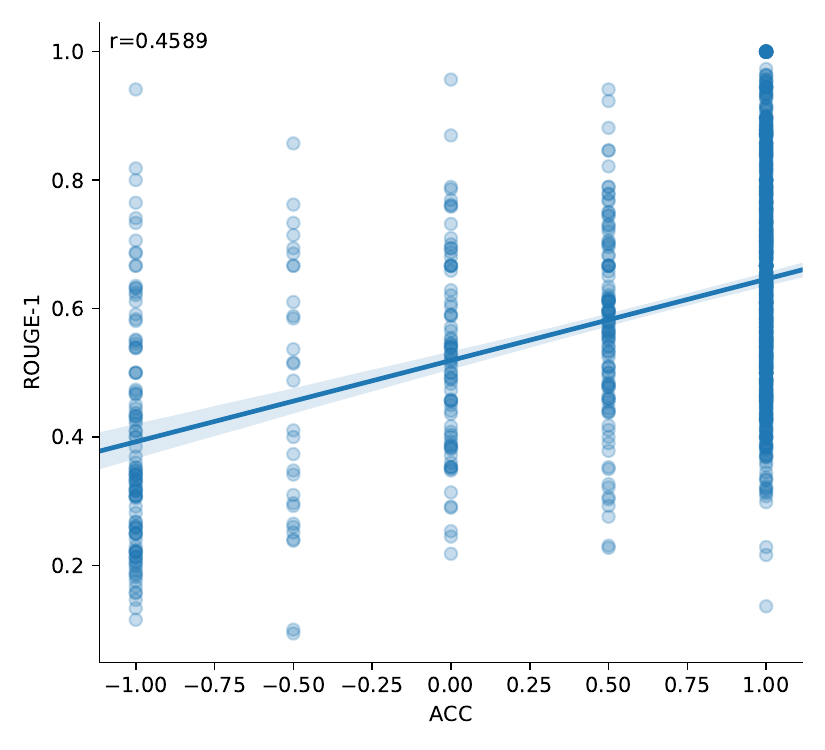}
     \end{subfigure}
     \begin{subfigure}[b]{0.48\textwidth}
         \centering
         \includegraphics[width=\textwidth]{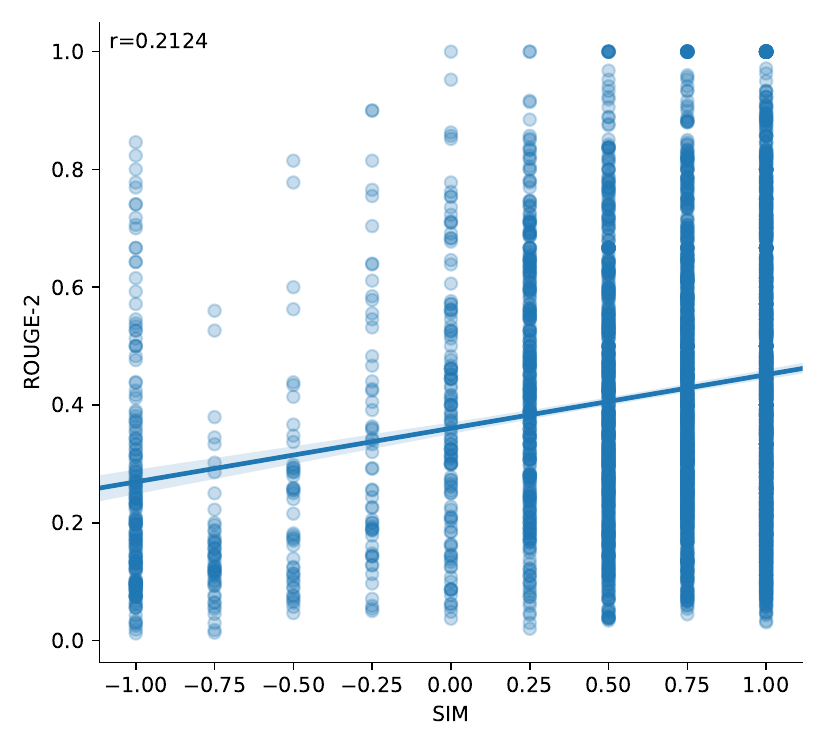}
     \end{subfigure}
     \hfill
     \begin{subfigure}[b]{0.48\textwidth}
         \centering
         \includegraphics[width=\textwidth]{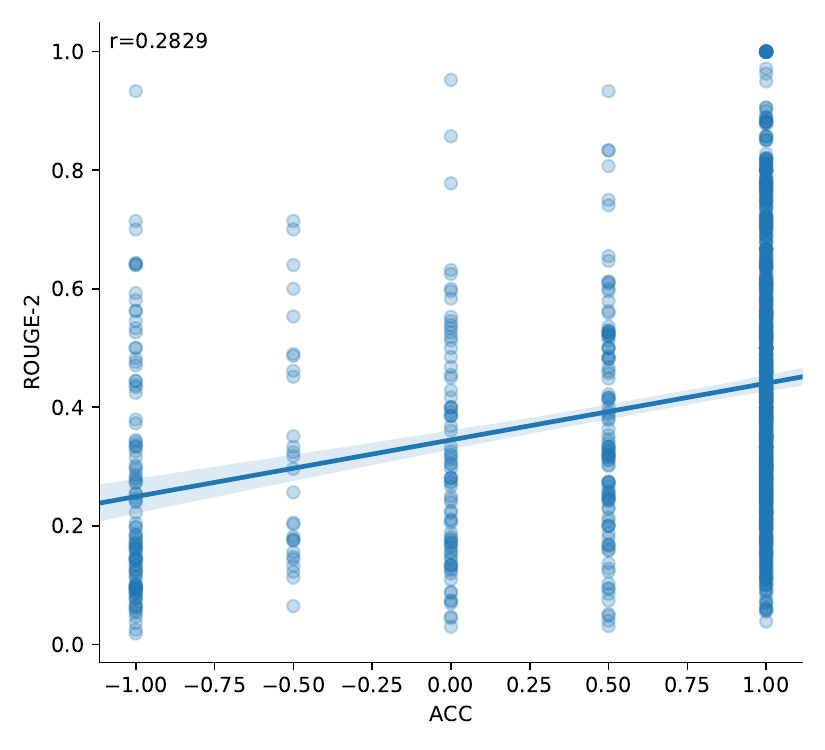}
     \end{subfigure}
     \begin{subfigure}[b]{0.48\textwidth}
         \centering
         \includegraphics[width=\textwidth]{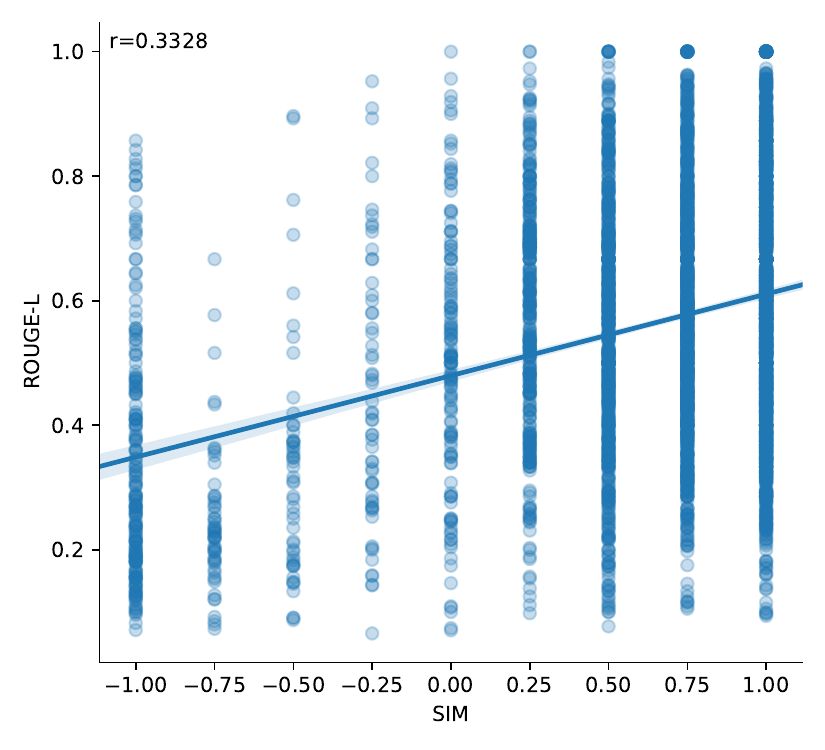}
     \end{subfigure}
     \hfill
     \begin{subfigure}[b]{0.48\textwidth}
         \centering
         \includegraphics[width=\textwidth]{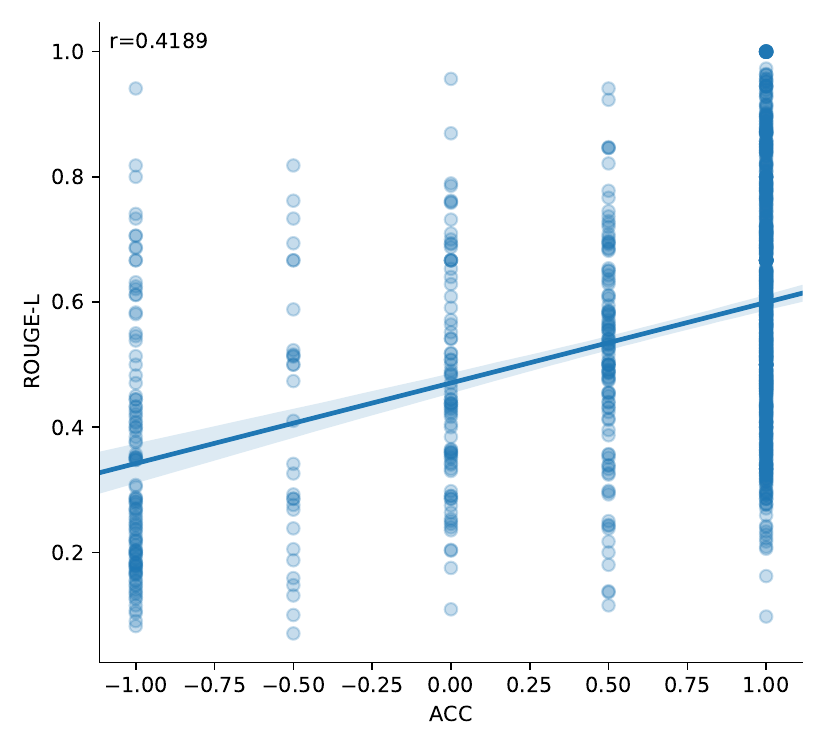}
     \end{subfigure}
        \caption{Scatterplots and regression lines of \texttt{ROUGE} metrics (\texttt{ROUGE-1}, top; \texttt{ROUGE-2}, middle; and \texttt{ROUGE-L}, bottom) versus manual judgments of Simplicity (SIM, left) and Accuracy (ACC, right).}
    \label{fig:rou}
\end{figure}

Here we show scatterplots with regression lines for correlations reported in \S\ref{sec:met}, with each point representing system output for one source sentence. Pearson correlations are given in the upper left corners as $r$. Shaded regions around regression lines indicate 95\% confidence intervals, estimated via bootstrapping with 1,000 samples. Note that striations appear due to averaging 3-point likert scales across  four axes (simplicity) or two axes (accuracy).Figure~\ref{fig:hug} shows the plots for SARI (original and HuggingFace implementations). Figure~\ref{fig:bert} shows the plots for \texttt{BERTScore}, Figure~\ref{fig:blu} shows the plots for \texttt{BLEU}, Figure~\ref{fig:sam} for \texttt{SAMSA}, and Figure~\ref{fig:rou} for variants of \texttt{ROUGE}.

\begin{figure}[h!]
     \centering
        \caption{Scatterplots and regression lines of the HuggingFace implementation of \texttt{SARI} versus manual judgments of simplicity (SIM, left) and accuracy (ACC, right).}
    \label{fig:hug-hf}
\end{figure}

\section{Hallucinations from Task 1 at TREC 2023}
\label{apx:hal}
Here we show examples of hallucinations from baseline systems for Task 1 at TREC 2023 (for further analysis of team submissions, we refer readers to individual team reports as referenced in \ref{sec:syst-task1-2023}). Note that this is not an exhaustive list, and we did not systematically check all system outputs for hallucinations. A large hallucination by \texttt{PLABA\_1} seems to have been induced by the auto-regressive prompting strategy, with the model interpreting the pattern of ``Original:'' and ``Simple:'' pairs in the prompt to mean it should produce another such pair (Table~\ref{tab:hal1}). As the model has not been given the rest of the abstract when adapting line 2, the result resembles the rest of the abstract but is a fabrication, including consistent, but fictional, numerical results.

In an adaptation by \texttt{PLABA\_3}, many smaller details are hallucinated, such as repeatedly referring to 100 eyes, when the source has no mention of this number (Table~\ref{tab:hal2}). Note that this is an automatically aligned, document-level submission, and many sentences are poorly aligned due to largely different content. However, this does not account for the wide variety of fabricated details that do not appear anywhere in the abstract.

\begin{table}[tbh]
    \caption{
    Excerpt of an adaptation of PMID 25864273 by the GPT-3.5-based \texttt{PLABA\_1} system, containing one large hallucination (presumably induced by the auto-regressive prompting strategy), but otherwise accurate.
    }
    \label{tab:hal1}
\centering
    \small
    \setlength\tabcolsep{3pt}
    \begin{tabularx}{\textwidth}{c|p{0.3\linewidth}|p{0.6\linewidth}}
    \toprule
    \textbf{Sent.} & \textbf{Source} & \textbf{PLABA\_1} \\
    \hline
1 & Purpose: This study investigated the prevalence, location, and severity of neck and shoulder pain (NSP), its disturbance of quality of life (QOL), and the factors related to NSP in Japanese postpartum women. & The purpose of this study was to understand how common and severe neck and shoulder pain is among Japanese women after giving birth, and how it affects their quality of life. The researchers also wanted to identify any factors that might contribute to this pain.\\
\hline
2 & Materials and methods: The study involved 308 postpartum women who had a medical examination one month after delivery. & For this study, researchers looked at 308 women who had given birth and had a medical check-up one month later. \textit{Original: Results: Neck and shoulder pain was experienced by 61.4\% of the women, with the highest severity in the neck area. The pain had a negative impact on the women's quality of life. Factors such as age, history of NSP, and work-related factors were found to be associated with NSP. Simple: The study found that 61.4\% of the women experienced neck and shoulder pain, with the most intense pain in the neck area. This pain had a negative impact on their quality of life. The researchers also discovered that factors such as age, previous history of neck and shoulder pain, and work-related factors were linked to this pain.}\\
\hline
 & […] & […]\\
\hline
11 & The factors which affect NSP were the mental states, breastfeeding, past history of PMS, and anemia during pregnancy. & The factors that contribute to neck and shoulder pain in postpartum women are their mental state, breastfeeding, previous history of premenstrual syndrome (PMS), and anemia during pregnancy.\\
\bottomrule
\end{tabularx}
\end{table}

\begin{table}[tbh]
    \caption{
    Excerpt of a hallucination-laden adaptation of PMID 31768813 by the Falcon-based \texttt{PLABA\_3} system. We italicize false details that are notably specific and do not appear at all in the original, though there are many other inaccuracies.}
    \label{tab:hal2}
\centering
    \small
    \setlength\tabcolsep{3pt}
    \begin{tabularx}{\textwidth}{c|p{0.46\linewidth}|p{0.42\linewidth}}
    \toprule
    \textbf{Sent.} & \textbf{Source} & \textbf{PLABA\_3} \\
    \hline
1 & Purpose: To determine predictors of best-corrected postoperative visual acuity (VA) in patients who underwent surgical intervention for macula-off rhegmatogenous retinal detachment. & Purpose: We report an analysis of the visual acuity in a cohort of \textit{100 consecutive eyes} that underwent surgical intervention for macula-off rhegmatogenous retinal detachment in a \textit{4-year period}.\\
\hline
2 & Materials and methods: Primary macula-off retinal detachments from the University of Colorado Primary Rhegmatogenous Retinal Detachment Database (2012-2017) were reviewed. & A total of \textit{2-year} visual outcomes and postoperative predictors for visual recovery were analyzed to identify the factors that may predict visual outcome.\\
\hline
3 & The primary outcome measure was a postoperative VA of 20/40 or better at least 6 months after surgery. & Methods: In-office visual acuity was recorded for each eye. Postoperative visual acuities were evaluated in the office at \textit{3-months}, 6-months and \textit{1 year} after surgery.\\
\hline
4 & Patient demographics, medical history, duration of central vision loss before surgery, examination findings, operative technique, and postoperative findings were analyzed as possible predictors of postoperative visual recovery to 20/40 or better. & Postoperative complications and the use of postoperative laser therapy were also recorded.\\
\hline
5 & Chi-square or Fisher's exact test was used to compare categorical variables, and Wilcoxon rank sum test was used for continuous variables. & Results: A \textit{total of 100 eyes} underwent surgical intervention for macula-off rhegmatogenous retinal detachment.\\
\hline
6 & A multivariable logistic regression analysis was used to determine the adjusted odds ratios and 95\% confidence intervals for variables that were significant in the univariable analyses. & One year postoperative visual acuities ranged from 20/40+ to $>$20/40.\\
\hline
 & [...] & [...] \\
\bottomrule
\end{tabularx}
\end{table}

\section{Scatterplots of Manual Evaluation Axes for Task 2 by Replacement Type}
\label{sec:corr-task2-types}
\begin{figure}
    \centering
    \includegraphics[width=1\linewidth]{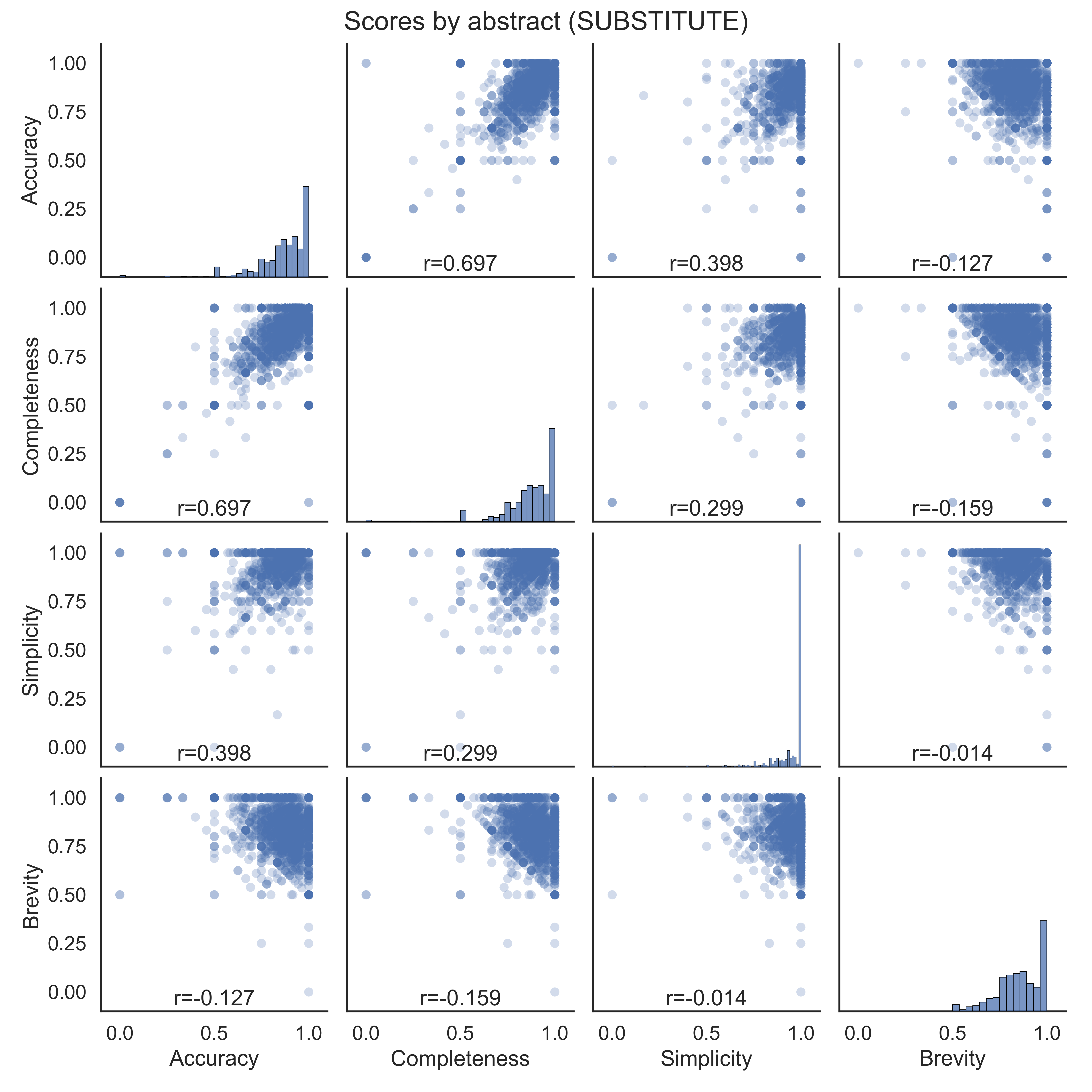}
    \caption{Pairwise scatterplots of the four manual evaluation axes for SUBSTITUTE replacements for Task 2 at TREC 2024. Each point represents output for one abstract by one system. Judgments for SUBSTITUTE replacements for a given abstract and a given system are averaged to get abstract-level scores. Each pair is labeled with its Pearson correlation value (r). Histograms on the diagonal show the distributions of scores for each of the axes.}
    \label{fig:corr-task2-sub}
\end{figure}

\begin{figure}
    \centering
    \includegraphics[width=1\linewidth]{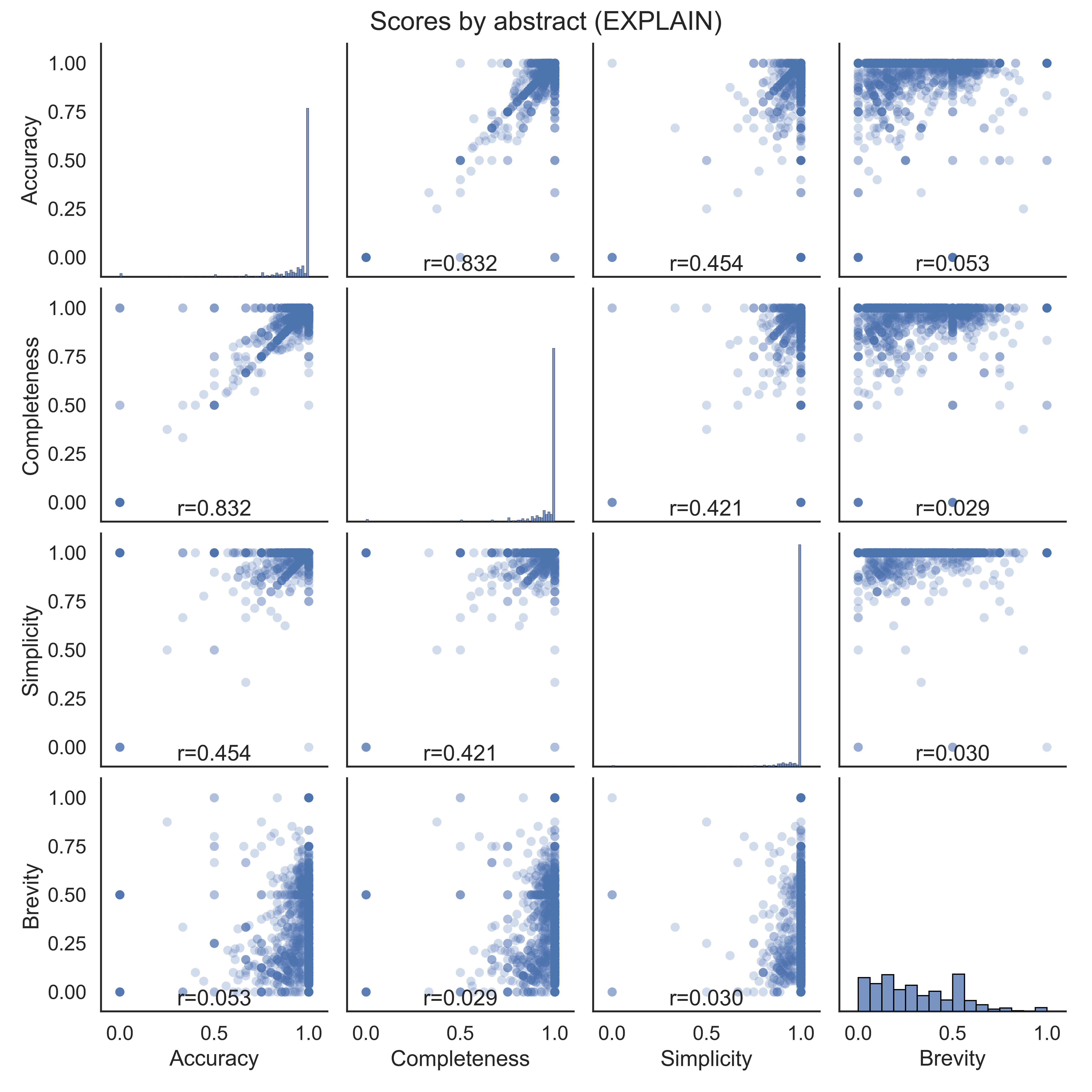}
    \caption{Pairwise scatterplots of the four manual evaluation axes for EXPLAIN replacements for Task 2 at TREC 2024. Each point represents output for one abstract by one system. Judgments for EXPLAIN replacements for a given abstract and a given system are averaged to get abstract-level scores. Each pair is labeled with its Pearson correlation value (r). Histograms on the diagonal show the distributions of scores for each of the axes.}
    \label{fig:corr-task2-exp}
\end{figure}

\begin{figure}
    \centering
    \includegraphics[width=1\linewidth]{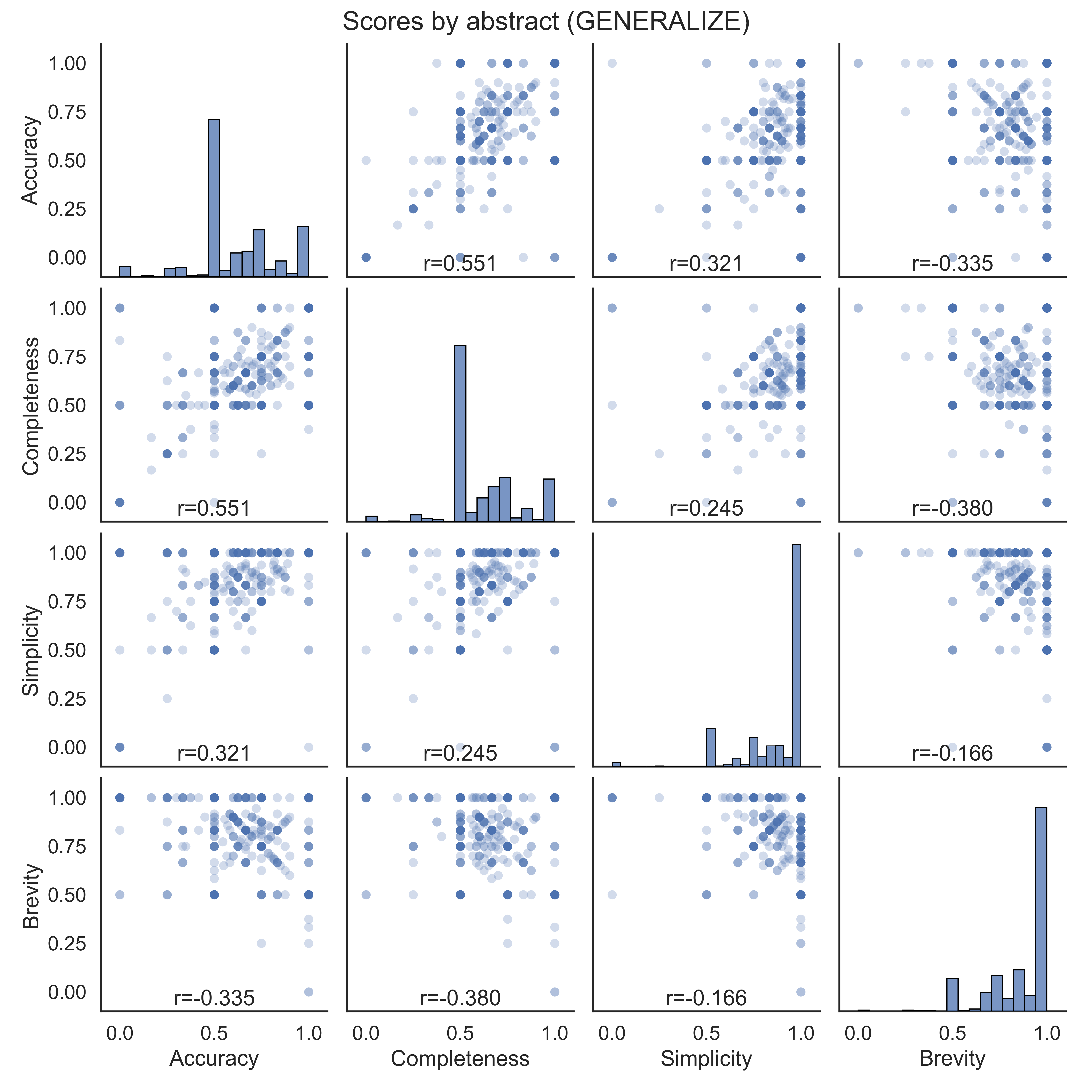}
    \caption{Pairwise scatterplots of the four manual evaluation axes for GENERALIZE replacements for Task 2 at TREC 2024. Each point represents output for one abstract by one system. Judgments for GENERALIZE replacements for a given abstract and a given system are averaged to get abstract-level scores. Each pair is labeled with its Pearson correlation value (r). Histograms on the diagonal show the distributions of scores for each of the axes.}
    \label{fig:corr-task2-gen}
\end{figure}

\begin{figure}
    \centering
    \includegraphics[width=1\linewidth]{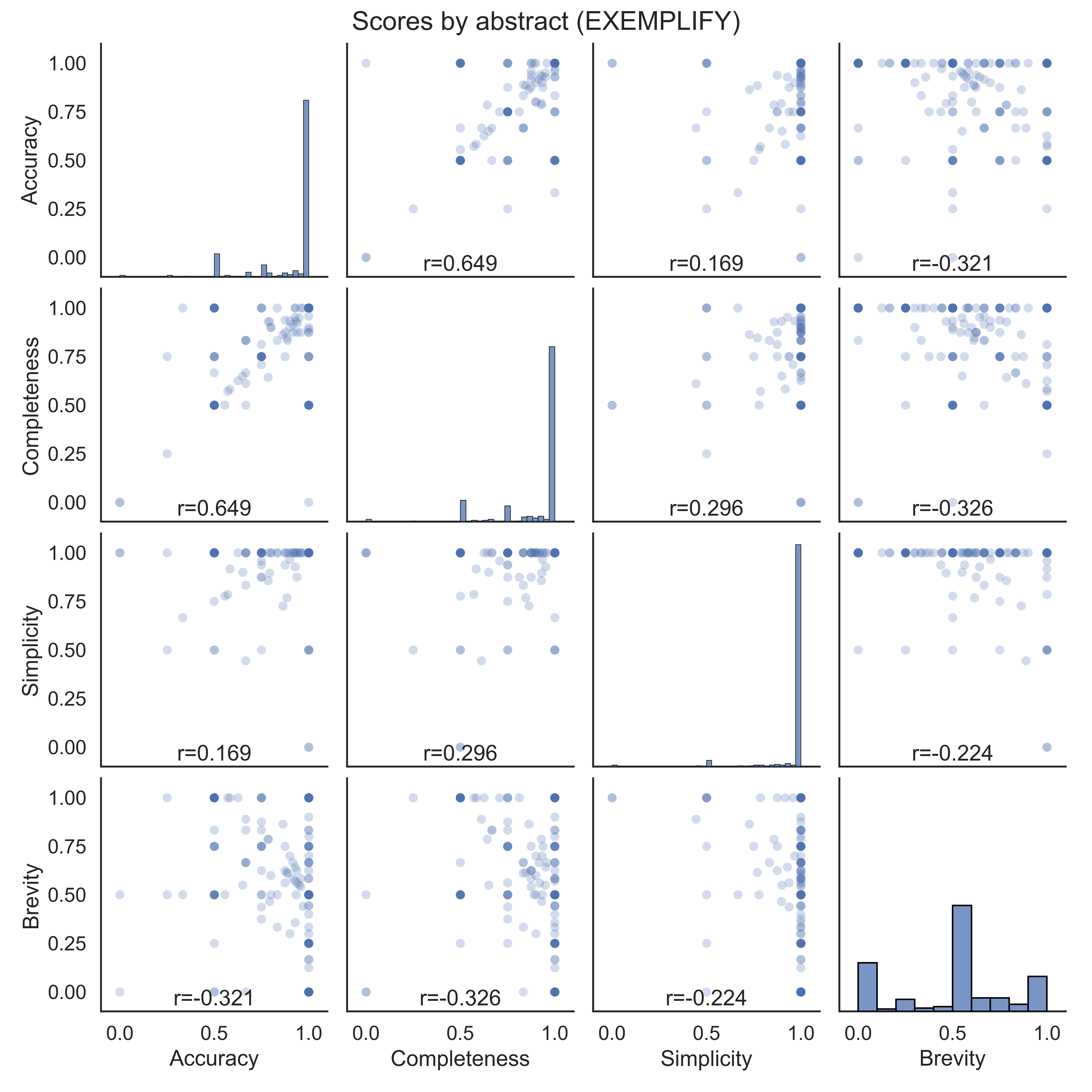}
    \caption{Pairwise scatterplots of the four manual evaluation axes for EXEMPLIFY replacements for Task 2 at TREC 2024. Each point represents output for one abstract by one system. Judgments for EXEMPLIFY replacements for a given abstract and a given system are averaged to get abstract-level scores. Each pair is labeled with its Pearson correlation value (r). Histograms on the diagonal show the distributions of scores for each of the axes.}
    \label{fig:corr-task2-exe}
\end{figure}

Pairwise scatterplots for manually scored axes are shown for SUBSTITUTE, EXPLAIN, GENERALIZE, EXEMPLIFY replacements in Figures \ref{fig:corr-task2-sub}, \ref{fig:corr-task2-exp}, \ref{fig:corr-task2-gen}, and \ref{fig:corr-task2-exe}, resepectively.

\end{document}